\documentclass[twocol]{ametsocV6}

\usepackage{hyperref}       
\usepackage{url}            
\usepackage{booktabs}       
\usepackage{amsfonts}       
\usepackage{nicefrac}       
\usepackage{cite}
\usepackage{amsmath}       
\usepackage{algorithm}
\usepackage{algpseudocode}
\usepackage{tikz}
\usetikzlibrary{matrix,calc}
\usepackage{caption}
\usepackage{subcaption}
\usepackage{cuted}

\renewcommand{\vec}[1]{\mathbf{\boldsymbol{#1}}}

\makeatletter
\renewcommand{\p@subsection}{\thesection\protect\eatbracket}
\makeatother
\def\eatbracket#1#2{#1\ifx)#2\else#2\fi}

\begin{document}

\title{Unpaired Downscaling of Fluid Flows with Diffusion Bridges}
%
%
\authors{Tobias Bischoff\correspondingauthor{tobias@caltech.edu} \&
Katherine Deck\thanks{Authors are listed in alphabetical order as co-first authors.}\thanks{All authors contributed equally to this work.}
}

\affiliation{
{Climate Modeling Alliance, California Institute of Technology, Pasadena, CA 91125, USA}
}

\abstract{We present a method to downscale idealized geophysical fluid simulations using generative models based on diffusion maps. By analyzing the Fourier spectra of images drawn from different data distributions, we show how one can chain together two independent conditional diffusion models for use in domain translation. The resulting transformation is a diffusion bridge between a low resolution and a high resolution dataset and allows for new sample generation of high-resolution images given specific low resolution features. The ability to generate new samples allows for the computation of any statistic of interest, without any additional calibration or training. Our unsupervised setup is also designed to downscale images without access to paired training data; this flexibility allows for the combination of multiple source and target domains without additional training. We demonstrate that the method enhances resolution and corrects context-dependent biases in geophysical fluid simulations, including in extreme events.  We anticipate that the same method can be used to downscale the output of climate simulations, including temperature and precipitation fields, without needing to train a new model for each application and providing a significant computational cost savings.}
 
\maketitle

\statement
The purpose of this study is to apply recent advances in generative machine learning technologies to obtain higher resolution geophysical fluid dynamics model output at lower cost compared with direct simulation while preserving important statistical properties of the high resolution data. This is important because while high-resolution climate model output is required by many applications, it is also computationally expensive to obtain. 

\section{Introduction}
Climate simulations are powerful tools for predicting and analyzing climate change scenarios, but they are often limited by computational resources and hence in their spatial and temporal resolution. As a result, simulations can both lack the high-resolution detail needed for many applications as well as carry an inherent bias due to the lack of representation of small-scale dynamical processes which feed back on larger scales. For example, 
horizontal resolutions of $O(10-100~\mathrm{km})$ are still too coarse to accurately simulate important phenomena such as convective precipitation, tropical cyclone dynamics, and local effects from topography and land cover, and hence these simulations can be of limited use for making predictions on regional and sub-regional scales. In particular, the low resolution fields suffer from biases in extreme temperatures and precipitation rates, which in turn can reduce the accuracy of projections of climate hazards on smaller spatial scales, e.g., \citet{abatzoglou2012comparison, gutmann2014intercomparison, hwang2014assessment}. 

Several approaches have been developed to address biases and increase resolution in fluid and climate simulations, a process referred to as ``downscaling" of fluid flows \citep{fowler2007linking, salathe2007review, maurer2008utility}. Nudging techniques, which involve constraining the solution of a dynamical system to follow the large-scale information, are applied during simulation-time and are a form of dynamical downscaling. On the other hand, statistical downscaling refers to methods which use a data-derived model to make a correction to a fluid simulation or to computed statistic after the simulation has been run, which can keep the computational budget comparatively lower. This is usually achieved by amortizing the training cost of these models over many evaluations during post-processing, as opposed to solving the highly resolved fluid system whenever output is needed (but not requiring any training time).

The predominant statistical method for bias correction and resolution enhancement of climate variables is the bias-correction spatial disaggregation (BCSD) algorithm \citep{panofsky1958some, wood2002long, wood2004hydrologic, thrasher2012bias}. BCSD uses quantile mapping to correct biases and Fourier transforms to enhance resolution. However, the BCSD algorithm has limitations, including its inability to incorporate auxiliary datasets and its lack of multivariate capability. Quantile mapping can also adversely affect large-scale features such as the evolution of mean values \citep{hagemann2011impact,pierce2013probabilistic, maurer2014bias, climagan}. Constructed analogs and variants thereof are a multivariate method for downscaling which have been shown to outperform other methods, perhaps due to the fact that they take into account correlations between variables \citep{pierce2014statistical,abatzoglou2012comparison}. However, these methods do not truly allow for sampling high resolution data from a distribution. 

At the same time, generative machine learning models like Generative Adversarial Networks (GANs; e.g., \citet{goodfellow2020generative}), Variational Autoencoders (VAEs; e.g. \citet{kingma2019introduction}), and Normalizing Flows (NFs; e.g., \citet{papamakarios2021normalizing}) have been demonstrated to be effective for super-resolution and for domain translation (e.g., \citet{wang2020deep, zhu2017unpaired}), and are now being applied to the downscaling task. In these use-cases, the models are deep convolutional neural networks which map images with multiple channels (e.g., fields of climate model variables at low resolution) into output images with multiple channels (e.g., fields of climate model variables at high resolution).  Statistical relationships between datasets are learned implicitly, either in a ``supervised" or ``unsupervised" fashion. In supervised approaches, data points are mapped from low resolution to high resolution in an aligned fashion: each low-resolution data point in the training set has a corresponding high-resolution data point in the training set.  Alternatively, when paired data points are unavailable, the downscaling task becomes one of modeling conditional data distributions, and then generating samples from these distributions. 

In this work, we propose to use generative models based on diffusion maps for generating downscaled fluid flows using unpaired training data.  Diffusion models have shown great flexibility in generating realistic samples from a variety of learned high-dimensional probability distributions (e.g., images, audio, and video, \citet{DhariwalNichol, kong2020diffwave, HoVideoDiffusion, ho2022imagen}). With respect to domain translation and downscaling, i.e., transforming a sample from a source distribution into a sample from a target distribution, generative diffusion models have distinct advantages over classical methods, e.g.  
\begin{itemize}
    \item generative models allow for sampling from high-dimensional probability distributions. From these samples any statistical quantity can be computed;
    \item diffusion-based models can be trained with unpaired data, and can therefore be used for multiple domain translation tasks without retraining for each source domain/target domain pair \citep{Su2022};
    \item pretrained diffusion-based models can be ``repurposed`` to sample from specific parts of the domain using guided sampling e.g., \citep{ClassifierFreeGuidance, DhariwalNichol}.
\end{itemize}
These points put diffusion-based models into a class distinct from classical methods and in some cases distinct even from GAN-based methods, many of which require paired data or paired source/target domains. This suggests diffusion-based models as a promising candidate well-suited for applications in fluid dynamics and climate science because
\begin{itemize}
    \item retraining machine learning models frequently is undesirable due to the potentially high training cost involving high dimensional data points (e.g., full climate fields);
    \item for downscaling tasks, paired datasets of high and low resolution climate {\it simulations} do not truly exist, due to deterministic chaos and the feedback of small scale motion to large scales;
    \item extreme events with biased tail probabilities can be correlated across climate variables and spatial locations, and calibrating a downscaling method for all statistics of interest is challenging. As a result the ability to generate samples can be highly desirable.
\end{itemize}
In this work, we provide a demonstration of how diffusion models can be used for domain translation between low and high resolution fluid simulations, without customization to the specific translation task under consideration. We focus on the generation of consistent high-resolution information given a low-resolution input and on the correction of important statistical biases, e.g., shifts in mean values, unresolved spatial scales, and underestimated tail events.

\subsection{Related Work}
\subsubsection{Downscaling of climate data}\label{sec:related_work_a}
Machine learning based methods for downscaling and bias correction have been applied to climate simulations successfully in prior works. \citet{pan2021learning} use GANs to bias correct climate simulation data over the continental United States and focus on matching various statistical quantities of corresponding observational data. Using paired high resolution radar measurements, ECMWF simulation data, and other contextual information, \citet{harris2022generative} use GANs to downscale simulated low resolution precipitation fields. They found that their model outperformed many conventional approaches, including on extreme rainfall events. Similarly, \citet{price2022} show that conditional GANs can be used to directly bias correct and downscale low resolution precipitation forecasts using high-resolution ground truth radar observations. \citet{climagan} use contrastic translation GANs and high resolution observations to downscale CMIP low resolution simulation data for daily maximum temperature and precipitation. This particular variant of GAN model allows for training in an unsupervised, unpaired fashion \citep{park2020contrastive}. Again, the authors find comparable or improved performance compared to existing methods. Similarly, \citet{climalign} use unpaired datasets to learn a domain translation map from low resolution simulation data to high resolution, unbiased data by combining normalizing flows with a cycle consistency loss function similar to that of CycleGANs \citep{zhu2017unpaired}. Methods based on CycleGANs generally require that a model is trained with access to both the source and the target data, so that a new model must be created for each translation task.

\subsubsection{Diffusion modeling}
Diffusion is a dynamical process  which erases initial conditions on long timescales. Using observed data (which are samples from an unknown data distribution) as initial conditions, we can integrate a trajectory forward in time under a diffusion model chosen so that as $t\rightarrow \infty$, the long-time steady state of the system corresponds to samples from a known distribution like a Gaussian (the prior distribution). Using samples from the prior distribution as initial conditions, solving the reverse-diffusion model will generate samples from the unconditional data distribution. That diffusive processes can then be used as generative models, transforming samples from a known prior distribution into samples from an unknown data distribution via a diffusive process, has been established by several authors, e.g., \citet{sohl2015deep, Song.2019, Ho2020}.  Indeed, more recently, \citet{ScoreBasedSDE_Song} showed that images created by generative diffusion models can be understood to be numerical solutions to ``reverse diffusion" stochastic differential equations, with initial conditions equal to samples from the prior distribution. This relies on the fact that (forward) diffusion processes can be reversed if the score, related to the gradient of the data distribution, is known \citep{ANDERSON1982}. Moreover, while the unconditional data distribution can be challenging to approximate directly, \citep{hyvarinen2005estimation,Vincent2011} have demonstrated how to approximate the score of the distribution using neural networks and gradient descent of a tractable loss function. 

The mathematical results for unconditional distributions, described above, can be extended to conditional distributions, allowing for conditional sampling \citep{ScoreBasedSDE_Song,Batzolis2021}. Many conditional diffusion models require paired input data points, and many important conditional generation tasks provide this type of data (e.g, super-resolution, inpainting, colorization, and other imputation tasks, including with temporal sequences - \citet{tashiro2021csdi, saharia2022image, Giannone, HoVideoDiffusion, Palette}). Alternatively, \citet{SDEdit} show how to generate photo-realistic images from simple stroke paintings with little detail by choosing an appropriate starting point (initial condition) and starting time for a reverse diffusion trajectory. As we will discuss, this can be interpreted as generating a high resolution image conditional on the input stroke painting. Crucially, it is carried out using a model trained only on the high resolution data, without access to the stroke paintings. That is, it does not require paired data to generate samples from the conditional distribution.

In constrast to CycleGANs \citep{zhu2017unpaired}, \citet{Su2022} show how diffusion models can be used for domain translation such that a model is trained once per data domain and not once per translation of interest (StarGANS are another solution, but at the expense of increased complexity \citet{choi2018stargan}). This feature of diffusion models arises because diffusion models for two domains have easily relatable prior distributions, and is advantageous because it allows for the same model to be used in many translation tasks. The translation works by completely diffusing an image from one domain, then turning that final state into a sample from the other domain's prior, and carrying out the reverse diffusion for the other domain using its model. The approach we will showcase in this work is based on combination of the ideas of \citet{SDEdit} and \citet{Su2022}, where a chain of diffusion models acts as a bridge between data domains, although the general idea has been around for much longer \citep{schrodinger1931}.

\subsection{Our Contribution}
We present and test an unsupervised method for the downscaling of fluid simulations. The method is based on chaining together diffusion-based generative models. It relies on the fact that coarsely resolved and highly resolved climate simulations differ on small and intermediate spatial scales, but mostly agree on the largest scales. Because the diffusion processes we employ here erase information on the smallest scales first, we can start with samples from a source domain (low-resolution), diffuse them until small-scale information is lost, then reverse diffuse them using a pre-trained diffusion model for the target domain (high-resolution). How well the resulting images match the source image on large scales while simultaneously containing fine scale features which match statistics of the high resolution data is governed by the time at which we stop the forward noising process and begin the reverse diffusion process. Following \citet{SDEdit} and \citet{Su2022}, we will refer to such a source-to-target diffusion model as a diffusion bridge. As described already, this approach has advantages over existing downscaling methods as it allows for sample generation, use with unpaired data, and the reusability of trained models, but it has not been tested yet for this application. 

Additionally, we introduce architectural improvements for the neural network employed in the diffusion model. These improvements are secondary to our overall goal but improve performance metrics and decrease training time. Score-based diffusion models are known to suffer from a ``color-shift": generated images may have the correct spatial features, but are shifted to different average colors relative to the training data.  The error grows for larger images. One approach for improving this artifact is to use an exponential-moving-average (EMA) of the model parameters, typically using a very long memory implied by the exponential moving average \citep{Song2020_ImprovedTechniques}. As a consequence, a large number of training iterations is required to reach good performance. While alleviation of the color-shift via other techniques is possible, e.g., \citet{Choi2022}, we reduce the effect by introducing a bypass layer in the neural network architecture. This removes the need for the exponential moving average even for large image sizes (e.g., 512x512 pixels in spatial resolution).

\section{Data \& Simulations}\label{sec:data_generating_model_main_text}
We use a two-dimensional advection-condensation model similar to the one proposed in \citet{ogorman_2006} to create the fluid simulation data used in this work. The model is intended to provide an approximate representation of the dynamics of moisture on isentropes in the extratropical atmosphere on Earth-like planets \citep{ogorman_2006}. As such it is an idealized toy model. Nevertheless, the model allows for detailed investigations of spectral and distributional properties of its vorticity and moisture fields at low computation cost compared to a full climate model. Throughout this study, we focus on two quantities with nearly isotropic statistics, vorticity and an advected tracer that represents the supersaturation $q'=q-q_\mathrm{s}$ in an Earth-like atmosphere, where $q$ is the mass fraction of water relative to moist air (specific humidity), and $q_s$ is the mass fraction when the air is saturated (saturation specific humidity). 

The vorticity evolves according to the two-dimensional Euler equations with random forcing and linear dissipation. The supersaturation is advected by the flow field implied by the vorticity field. It is forced by a spatially homogeneous source $e$ that can be interpreted as an evaporation field, adding moisture to the flow, as well as a spatially varying condensation which decreases moisture in situations of supersaturation $q > q_\mathrm{s} (q' > 0)$. Condensation therefore represents the tail of the supersaturation tracer distribution, and extreme condensation events are correspondingly even further into the tail. For more mathematical details on the idealized advection-condensation model, see Appendix~\ref{sec:FluidModel}.

\begin{figure}[htbp]
\centering
\includegraphics[width=0.5\textwidth]{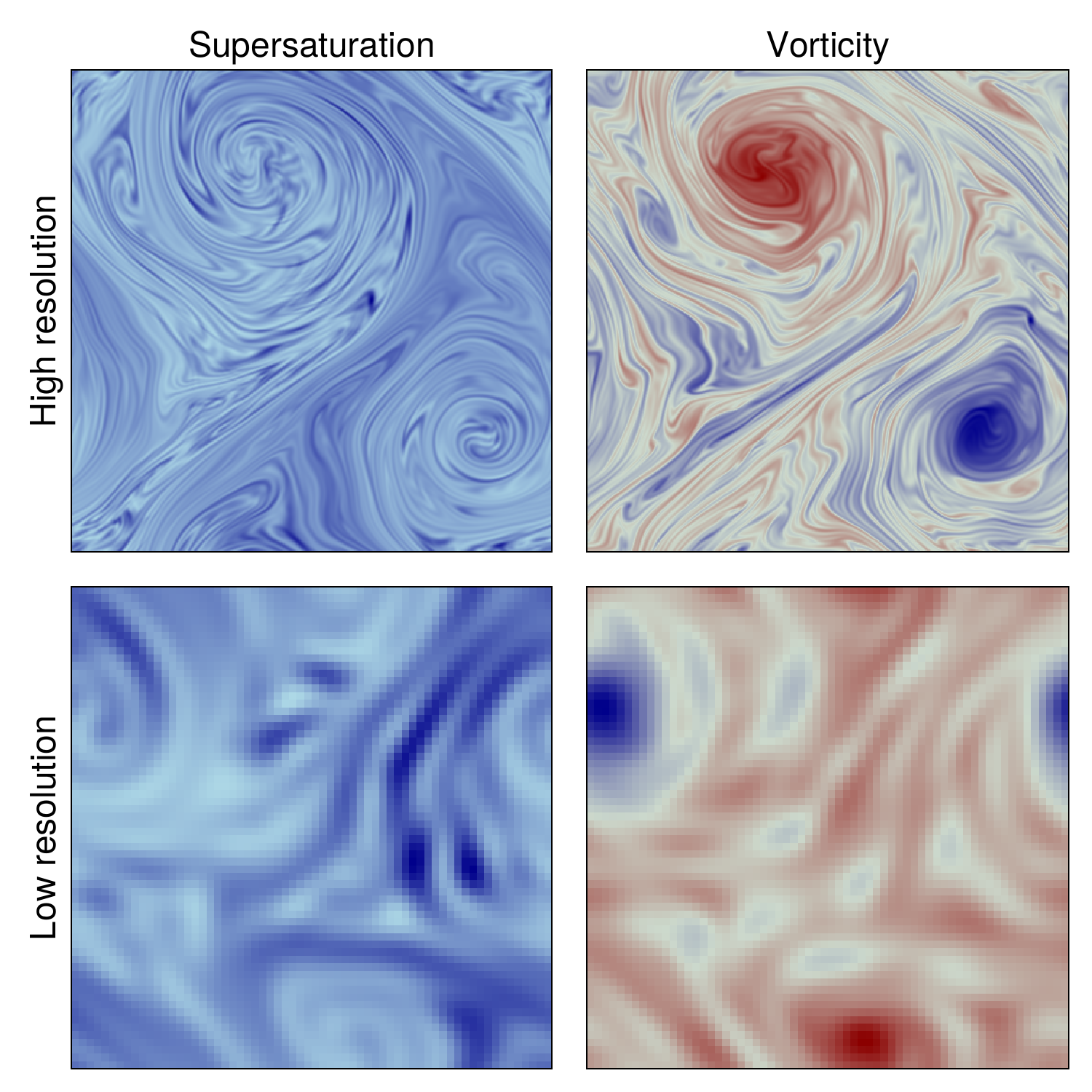}
\caption{Random snapshots from the two-dimensional fluid dynamics models. The top row shows high resolution snapshots, while the bottom row shows low resolution snapshots. The left column shows the \textbf{supersaturation tracer} field and the right column shows the \textbf{vorticity} field. The values corresponding to the colors are irrelevant for the methods presented in this paper, but for the supersaturation field positive values (blue colors) correspond to regions in which the condensation term in the simulation model is active (e.g., idealized rainfall events occur). The white regions in the supersaturation field are areas of saturation deficits (no idealized rainfall events). For the vorticity field red colors correspond to positive values (cyclonic vorticity), while blue values correspond to negative values (anticyclonic vorticity).}
\label{snapshots}
\end{figure}

In order to mimic the meridional decay of the saturation specific humidity $q_s$ along isentropes in Earth's atmosphere, we assume a linearly decaying profile that is modulated by a spatially periodic perturbation. The spatially periodic perturbation is useful because it can be used to impose spatial inhomogeneities in supersaturation tracer statistics at different lengths scales. Loosely speaking, these inhomogeneities can be interpreted as a very idealized version of orographic impact on the saturation specific humidity fields.  

The time-independent form of $q_s$ is given by the following expression
\begin{align}
\label{eq:sat_moisture}
    q_s(x,y) = \gamma y + A \sin \left(\frac{2\pi k_x x}{L}\right) \sin \left(\frac{2\pi k_y y}{L}\right), 
\end{align}
where $\gamma$ denotes a background saturation specific humidity gradient, $A$ is the modulation amplitude, $L$ is the domain size in $x$ and $y$ direction, and $k_x = k_y$ denote wavenumbers that takes values $k_{x,y} \in \{1, 2, 4, 8, 16\}$ that allows for different large-scale saturation specific humidity profiles. As such, we can generate a dataset of supersaturation tracer fields with different idealized orographic or supersaturation tracer forcings. Our goal is to understand how well the diffusion model can make use of {\it contextual information} when downscaling, as high frequency variations in topography, surface coverage, and other fields affect the atmospheric flow in a more realistic climate simulation. This site-specific information is often included in generative models, including the ones described in Section \ref{sec:related_work_a}. Training the model with context will in principle lead to better performance and to better generalization. 

In order to generate data for training the denoising diffusion model, we performed a series of simulations at different resolutions and with different saturation specific humidity profiles, varying only the wavenumbers $k_x, k_y$ of the saturation specific humidity modulation. We generated a set of high resolution ($512 \times 512$ pixels) with varying background saturation specific humidity profile and a low resolution dataset ($64 \times 64$ pixels) with a fixed and unmodulated background saturation specific humidity field, i.e., $A=0$. The parameters used in the simulations are given in Table~\ref{tab:parameters}. Snapshots of these simulations once steady-state was reached were saved and used as training data for the diffusion model. We trained the diffusion model on the entire high-resolution dataset, including all context wavenumbers. Examples from the two-dimensional fluid dynamics models are shown in Figure~\ref{snapshots}. The top row shows high resolution snapshots, while the bottom row shows low resolution snapshots. The left column shows the supersaturation tracer field and the right columns shows the vorticity field, respectively. Finally, we resized the low-resolution $64 \times 64$ images using nearest-neighbor weighting to a resolution of $512 \times 512$, and removed high frequency aliases by applying a low pass filter. We ensured that the spectral information did not change between the true $64 \times 64$ images and our resized ones. 
 
\section{Downscaling with Diffusion Bridges}
\subsection{Diffusion Models}
 Our implementation of score-based generative models follows that of \citet{ScoreBasedSDE_Song}. The forward diffusion (``noising") process involves adding independent samples of Gaussian noise to each pixel, where the added noise has a mean of zero and a variance that depends on time in a prescribed fashion. Concretely, given an initial condition $\mathbf{x}(t=0) \sim p_{\mathrm{data}}(\mathbf{x})$ drawn from the data distribution, the noising process is defined by the stochastic differential equation (SDE)
\begin{align}\label{eq:forward_sde}
    \mathrm{d}\mathbf{x} = g(t) \mathrm{d}\mathbf{W},
\end{align}
 where $g(t)$ is a non-negative prescribed function of time and $\mathrm{d}\mathbf{W}$ implies a Wiener process. At any time $t$, the solution to this SDE is the ``noised" image $\mathbf{x}(t)$, which is drawn from a Normal distribution
\begin{equation}\label{eq:p_xt_given_x0}
\vec{x}(t) \sim \mathcal{N}(\vec{x}(0),\sigma^2(t)) = p(\vec{x}(t) | \vec{x}(0)), 
\end{equation}
where $\sigma(t)^2$ is the variance defined by 
\begin{equation}
\sigma^2(t) = \int_0^{t} g^2(t') \, \mathrm{d}t'.
\end{equation}
Here, we have chosen $g(t)$ such that at $t=1$, the variance $\sigma^2(t=1)$ is much larger in magnitude than the original pixel values, and hence all memory of initial conditions is lost, i.e.,
\begin{equation}
    p(\vec{x}(1) | \vec{x}(0)) \approx \mathcal{N}(\mathbf{0}, \sigma^2(1)).
\end{equation}
In this view, diffusion is a process which embeds a source image into a latent space, such that samples in the latent space $\vec{x}(1)$ are drawn approximately from a known distribution - which is independent of the source data. 

\begin{figure}[htbp]
\centering
\includegraphics[width=0.5\textwidth]{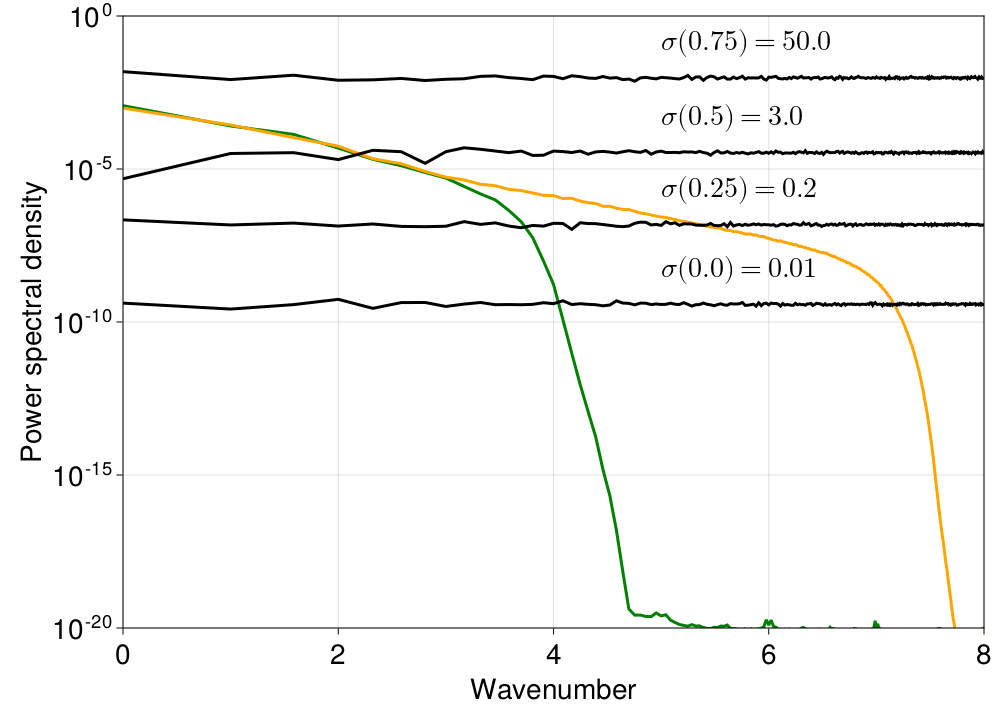}
\caption{The power spectral density for the vorticity field for both the low and high resolution datasets, along with the power spectral density of Gaussian white noise of different variance $\sigma(t)^2$. As larger amplitude Gaussian noise is added at larger diffusion times ($t\rightarrow 1$), the noising process will erase the information on the smallest scales first; see also \citet{Rissanen2022}.}
\label{bridge_graphic_a}
\end{figure}  

In order to approximately sample from the data distribution, we reverse this process. First, we sample from the latent-space prior distribution to obtain $\mathbf{x}(1)$. This is the initial condition for the reverse-diffusion equation, which is solved from $t=1$ to $t=0$.  The equation which reverses Equation \eqref{eq:forward_sde} is given by \citet{ANDERSON1982} as
\begin{align}\label{eq:reverse_sde}
\mathrm{d}\vec{x} & = -g(t)^2 \vec{s}(\vec{x}, t) \mathrm{d}t + g(t) \mathrm{d}\vec{W},
\end{align}
where $\vec{s}(\vec{x},t)$ is the score of the data distribution,
\begin{equation}
\vec{s}(\vec{x},t) \equiv  \nabla_x \log{p_{\mathrm{data}}(\vec{x})}.
\end{equation}
The goal of the training process used in diffusion modeling is to determine a parameterized representation of the score $\vec{s}_\theta(\vec{x}, t) \approx s(\vec{x}, t)$ through gradient descent on an appropriately chosen loss function. That is, we parameterize the time derivative appearing in the reverse SDE and learn it from the data. The final image $\mathbf{x}(t=0)$ resulting from this reverse simulation, with a trained model for the score, is the new data sample.

\subsection{Diffusion Bridges using Spectral Information}
Using a diffusion bridge for domain translation entails chaining together the forward model for a source data set, and the reverse model for a target data set (e.g., \citet{SDEdit, Su2022}). Our downscaling procedure builds on this idea; the discussion below makes it explicit why the approach works by making the connection to differential noising of spatial scales.
      
According to Equation \eqref{eq:p_xt_given_x0}, the noised image $\mathbf{x}(t)$ is the sum of $\mathbf{x}(0)$ and Gaussian noise, and so we can write its Fourier transform as the sum of the Fourier transform of $\mathbf{x}(0)$ and Gaussian noise (as the Fourier transform of Gaussian noise is (complex-valued) Gaussian noise). As these are uncorrelated, we can approximate the power spectral density of $\mathrm{PSD}_{\vec{x}(t)}(k)$, where $k = \sqrt{k_x^2 + k_y^2}$ is the wavenumber, as 
\begin{subequations}
\begin{align}
    \mathrm{PSD}_{\vec{x}(t)}(k) &\approx \mathrm{PSD}_{\vec{x}(0)}(k) + \mathrm{PSD}_{\vec{\eta}(t)}(k) \\
    \mathbf{\eta(t)} &\sim \mathcal{N}(\vec{0}, \sigma^2(t)).
    \end{align}
\end{subequations}
The power spectral density of white noise is independent of wavenumber: $\mathrm{PSD}_{\vec{\eta}(t)}(k) = \sigma(t)^2/N^2$, where $N$ is the image size. For reference, the power spectral density compresses the 2D Fourier transform into a 1d signal which is independent of direction. More details are provided in  Appendix~\ref{sec:PSD}. Snapshots of fluid flows generally exhibit a decay in power with increasing $k$. Hence, as the diffusion time increases from $t=0$ to $t=1$, and $\sigma(t)$ increases, the smallest scales (largest $k$) are noised first, cf. Figure~\ref{bridge_graphic_a} (see also \citet{Choi2022,Rissanen2022}).  

We assume the existence of a spatial scale $\lambda^\star$ above which the low resolution data is unbiased; a high resolution simulation passed through a low-pass filter would agree with the low resolution simulation for $\lambda > \lambda^\star$. The existence of $\lambda^\star$  implies that the expected power spectral density $\mathrm{PSD}(k)$ of the low resolution data and the high resolution data agree for wavenumbers $k < k^\star$, where $k^\star=2\pi/\lambda^\star$. Given the value of $k^\star$, one can therefore estimate  the diffusion time $t^\star$ at which signals on all spatial scales smaller than $\lambda^\star$ have a signal to noise ratio of $\lesssim 1$,
\begin{equation}\label{eq:t_star}
    t^\star =  \sigma^{-1}\bigg(\sqrt{N^2 \mathrm{PSD}(k^\star)}\bigg),
\end{equation}
since $\sigma(t)$ is a known analytic function. 

We are interested in translating an image from a source domain $\vec{x}_\mathcal{S} \in \mathcal{S}$ into an image from a target domain $\vec{x}_\mathcal{T} \in \mathcal{T}$. More concretely, our samples from $\mathcal{T}$ are $512 \times 512$ images generated by solving a fluid simulation at high-resolution, and our samples from $\mathcal{S}$ are $512 \times 512$ images generated by solving a fluid simulation at $64 \times 64$ resolution and then upsampling and low-pass filtering. All data $\vec{x}$ lie in $\mathbb{R}^{512\mathrm{x}512}$; by ``target" and ``source" domains $\mathcal{S}$ and $\mathcal{T}$, we refer to the lower dimensional manifolds within $\mathbb{R}^{512\mathrm{x}512}$ that we assume the data lie on\footnote{That trajectories generated by fluid simulations are constrained to lower dimensional manifolds seems plausible given the conserved quantities and partial differential equations governing the flow, and many large dimensional datasets are observed to lie on lower dimensional manifolds, e.g. \citet{brownverifying}.}. The downscaling (domain translation) algorithm defines a function mapping from $\mathcal{S}\in \mathbb{R}^{512\mathrm{x}512}$ to $\mathcal{T}\in \mathbb{R}^{512\mathrm{x}512}$. We use a sample $\vec{x}_\mathcal{S}$ as an initial condition and solve the forward noising model of the source domain to time $t^\star$. We then use $\vec{x}(t^\star)$ as an initial condition, and solve the reverse denoising model of the target domain to $t=0$. The resulting image $\vec{x}(0)$ is the generated image from $\mathcal{T}$. This transport map is probabilistic because different evaluations yield different samples from the target domain. That this process approximately samples from the conditional $p(\vec{x}_\mathcal{T} | \vec{x}_\mathcal{S})$ is not proven here, and may not be exact, but it is intuitive: the large scale features between the two images are kept fixed during this sampling process. The downscaling algorithm is defined more precisely in Algorithm \ref{alg:algorithm1}.

\begin{algorithm}
\caption{Downscaling Algorithm. Steps 1-3 only are carried out once, while (4-7) are carried out for each downscaled image.}\label{alg:algorithm1}
\begin{algorithmic}[1]
\State Compute the expected power spectral densities for the source and target domains, $\mathrm{PSD}_\mathcal{S}(k)$ and $\mathrm{PSD}_\mathcal{T}(k)$.
\State Solve for $k^\star$ such that  $\mathrm{PSD}_\mathcal{S}(k^\star) = \mathrm{PSD}_\mathcal{T}(k^\star) \equiv \mathrm{PSD}^\star$.  
\State Compute $t^\star(\mathrm{PSD}^\star)$, Equation \eqref{eq:t_star}.
\State Sample $\vec{x}_\mathcal{S} \sim p_{\mathrm{data},\mathcal{S}}(\vec{x})$
\State Obtain $\vec{x}(t^\star)$ by solving Equation \eqref{eq:forward_sde} from $t=0$ to $t= t^\star$, using $\vec{x}_\mathcal{S}$ as an initial condition.
\State Obtain  $\vec{x}(0)$ by solving Equation \eqref{eq:reverse_sde}, with $\vec{s}_\mathcal{T}(\vec{x}, t)$ as the score, from $t= t^\star$ to $t=0$, using $\vec{x}(t^\star)$ as an initial condition.
\State Return $\vec{x}(0)=\vec{x}_\mathcal{T} \sim p(\vec{x}_{\mathcal{T} | \vec{x}_\mathcal{S})}$.
\end{algorithmic}
\end{algorithm}

\begin{figure*}[htbp]
\centering
\includegraphics[width=\textwidth]{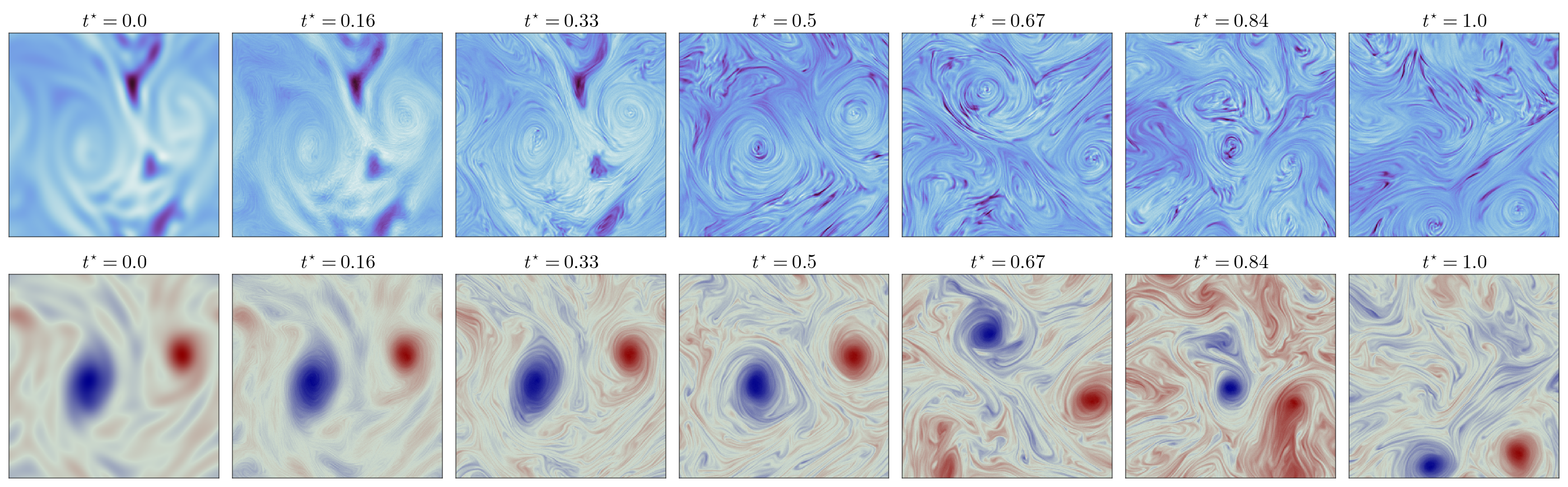}
\caption{The effect of $t^\star$ on the generated downscaled images of the \textbf{supersaturation tracer} (top row) and \textbf{vorticity} (bottom row). Too small of a value does not result in realistic looking high-resolution images, while too large of a value leads to realistic, but randomly chosen, high-resolution images. An optimal value of $t^\star\lesssim0.5$ yields a realistic downscaled version of the low resolution source image ($t=0$). For the supersaturation field positive values (blue colors) correspond to regions in which the condensation term in the simulation model is active (e.g., idealized rainfall events occur). The white regions in the supersaturation field are areas of saturation deficits (no idealized rainfall events). For the vorticity field red colors correspond to positive values (cyclonic vorticity), while blue values correspond to negative values (anticyclonic vorticity).}
\label{bridge_graphic_b}
\end{figure*}

Figure \ref{bridge_graphic_b} shows the generated images resulting from the downscaling procedure for different values of $t^\star$. For $t^\star \lesssim 0.5$, the diffusion bridge has preserved the large scale features of the low resolution image, but only the finest scale high resolution features have been filled in. Intermediate scales are missing. For $t^\star \gtrsim 0.5$, the forward noising process has erased some or all of the large scale features we wish to preserve. In the limit of $t^\star=1$, we have sampled a high-resolution image from $p_{\mathrm{data}, \mathcal{T}}$ without any information from the source image preserved. The optimal value, with respect to downscaling, is near $t^\star=0.5$, as also shown in Figure \ref{bridge_graphic_a}.

The core idea of finding an optimal value of $t^\star$ was also explored in \citet{SDEdit}, but without explaining the connection to spatial scales. In that work, they used a tradeoff between faithfulness to the ``guide" image (equivalent to our low resolution image) and realism with respect to the target image, and employed Kernel Inception Scores and L2 norms. Choosing the optimal $t^\star$ will rely on the metrics of interest to a given problem, and many methods may work well. 

It is important to note that this process requires that the power spectral density of the images of interest decrease with wavenumber. Additionally, our current implementation relies on the fact that different channels of the image have similar spectral density shapes, so that a single value of $t^\star$ works for each channel. A more general model would allow the different channels to have different schedules $\sigma(t)$.

\subsection{Contextual Information}\label{sec:context_main_text}
In the case of real climate simulations and data, the transport map may also depend on contextual information, such as surface properties, topography, etc. In order to study this, we employ a contextual field which our data generating model, an advection-condensation model, depends on (the dependence is explained in Section \ref{sec:data_generating_model_main_text} and Appendix \ref{sec:FluidModel}). The contextual information, which we denote generically as $\mathbf{x}_\mathcal{C}$, is aligned with the fluid state variable fields, denoted $\mathbf{x}$. As such, we are able to sample pairs from the distribution $(\mathbf{x}, \mathbf{x}_\mathcal{C}) \sim p(\mathbf{x}, \mathbf{x}_\mathcal{C})$. For the low resolution runs, the context is taken to be flat. At high-resolution, the context is available at the same resolution as the fluid state variable fields, and is spatially varying. In this work, we treat the context as an additional channel as input into the convolutional neural network modeling the score function. We do not carry out any ``diffusion" on these channels. More details on this can be found in Appendix \ref{sec:contextual_diffusion}.

Including contextual information does not change the explanation given above with respect to the diffusion bridge.  We assume the existence of a spatial scale above which the low resolution simulation is unbiased and above which contextual information has not affected the flow. A data point from the source (low resolution) domain is noised via Gaussian noise until the small-scale information is lost while the very largest scales remain approximately the same. Reverse diffusion is then applied to map the noised image towards the target (high resolution) domain. It is in the reverse diffusion process where contextual information enters and plays a role. This means that depending on the contextual information a different segment of the target domain is reached. This highlights how contextual information can be used to guide the generative process.

\begin{figure*}[htbp]
\centering
\includegraphics[width=\textwidth]{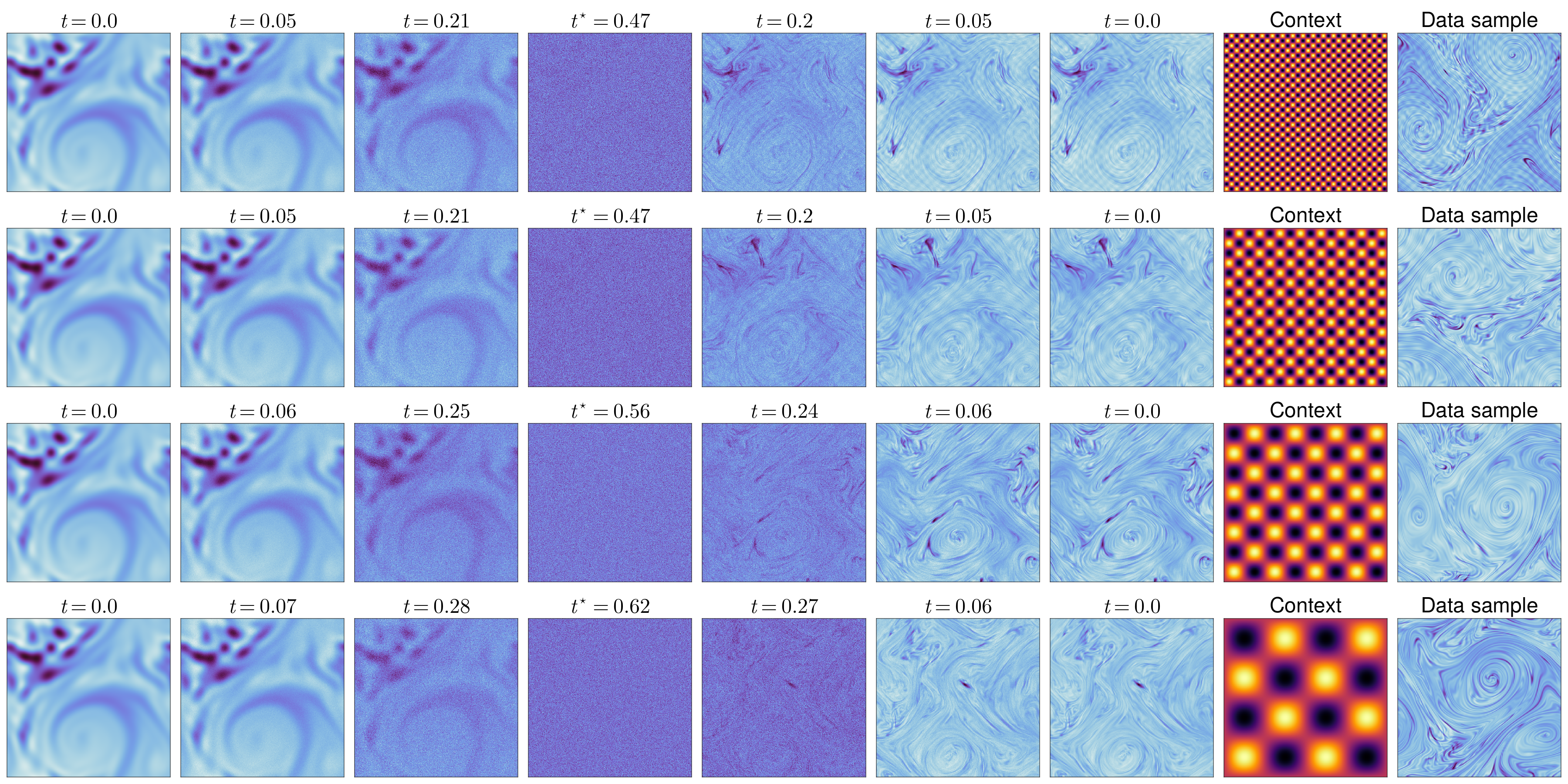}
\caption{Downscaling the same low resolution \textbf{supersaturation tracer} field, using four different contexts (by row). On the left is the source image, at $t=0$. Progressing to the right is equivalent to progressing through the downscaling procedure: the forward model of the source domain is used to noise the images, and at $t= t^\star$, we switch to the reverse model to integrate back to $t=0$. The generated high resolution samples (at $t=0$) have different periodic signals in the fluid flow; these are due to their specific context (second to rightmost column).  A randomly chosen data sample for each context is also shown (rightmost column). All noised images have been scaled to have the same range, which is necessary as the variance of the added noise grows with time. For the supersaturation field positive values (blue colors) correspond to regions in which the condensation term in the simulation model is active (e.g., idealized rainfall events occur). The white regions in the supersaturation field are areas of saturation deficits (no idealized rainfall events).}
\label{downscale_context}
\end{figure*}

The process is illustrated in Figure~\ref{downscale_context}, where we use the same source image to generate downscaled images with different contexts (we only show the supersaturation field; vorticity is unaffected by the context in our setup). For the $k_x=k_y = (8,16)$ cases (top two rows), the same value of $t^\star$ can be used because the high-resolution and low resolution power spectral densities only agree on spatial scales larger than the spatial scale of the contextual perturbation. On the other hand, for the $k_x=k_y=2$ case (bottom row), we must use a larger value of $t^\star$ in order to recover the signal from the contextual field. At the same time, this value of $t^\star$ is in the regime where low resolution information is being lost (as shown in Figure \ref{bridge_graphic_b}). The wavenumber $k_x=k_y=4$ case is in between.  The large scale features of the source image are present but distorted, and the modulation is less obvious than in the randomly drawn data sample with the same context. In Section \ref{sec:results}, we show the power spectral densities for the low and high resolution data sets, making this discussion more quantitative.

With respect to Figure \ref{downscale_context}, note that (1) even if each downscaling simulation used the same contextual fields, the generated high resolution images would be different due to the probabilistic nature of the downscaling process, and that (2) there is a single diffusion model used to generate the samples with different contexts. All data across all contexts was used to train this model.

\subsection{A Bypass for Spatial Mean Bias Reduction}
\label{sec:color_shift_main_text}

As discussed in the introduction, diffusion models can struggle to produce images with correct spatial means (``color shifts" in RGB images) while producing realistic spatial variations (e.g., power spectra appear reasonable). The recommended solution to this is to employ an exponential moving average (EMA) of the parameters of the model with a long memory \citep{Song2020_ImprovedTechniques}. In some score network architectures attention blocks are used (e.g. \citet{Ho2020}) which may also improve the colorshift, as self-attention allows for learning non-local features \citep{nonlocalnn}. Most of these approaches incur additional computational cost during training, which are not necessarily prohibitive, but may nevertheless be avoided.

\begin{figure*}[htbp]
\centering
\noindent\includegraphics[width=\textwidth]{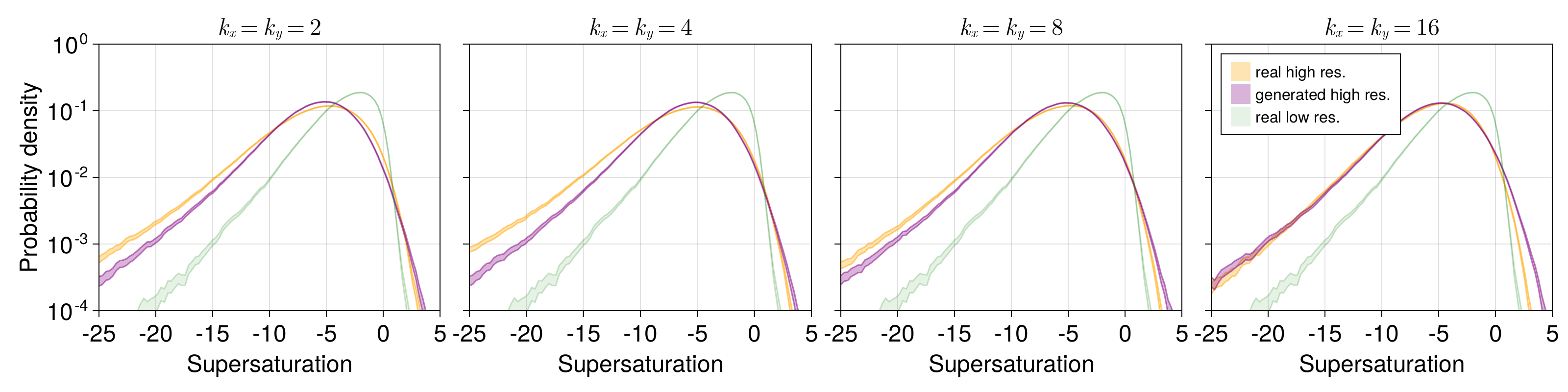}
\caption{Probability density function estimates for values of the \textbf{supersaturation tracer} field. Columns correspond to different high resolution data subsets, where $k$ indicates the saturation specific humidity modulation wavenumber. Orange distributions show distributions of real high resolution samples, purple distributions show distributions of generated and downscaled high resolution samples, and green distributions show distributions of real low resolution samples. Estimates of probability densities are performed via kernel density estimation. Shaded areas are computed from $10000$ bootstrap samples at the $99~\%$ confidence interval.}
\label{pdfs_ch1}
\end{figure*}
\begin{figure*}[htbp]
\centering
\noindent\includegraphics[width=\textwidth]{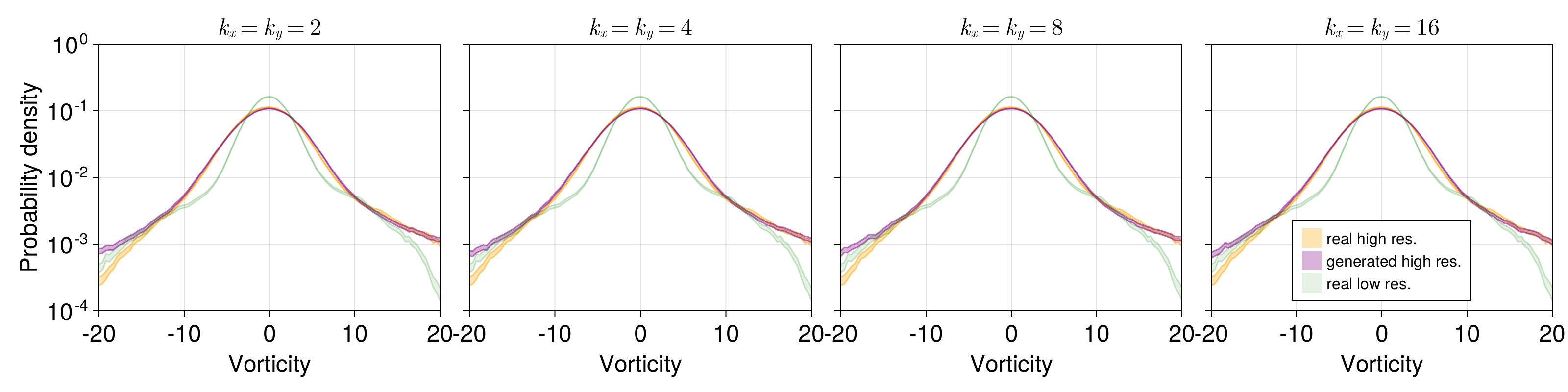}
\caption{Same as Figure~\ref{pdfs_ch1} but we are now plotting all quantities for \textbf{vorticity}. Unlike for the supersaturation tracer field values, vorticity field values are distributed nearly symmetrical around zero because the vorticity forcing in the simulations also has this property, wheras this is not the case for the supersaturation tracer.}
\label{pdfs_ch2}
\end{figure*}
Errors in the spatial means of images can only result from errors in the spatial mean of the score.  Though the neural networks used in score modeling have the capability to predict this, they do not learn to do so efficiently. A practical solution is to split the network's task into two individual, and independent, tasks: predicting the spatial mean of the score, and predicting the spatial variation about the mean of the score. We realize this by predicting the spatial mean of the score in a bypass layer of our network with independent parameters.  This allows us to keep our neural network architecture simple, essentially consisting only of a basic U-net \citep{unet}, and does not require the longer training process required by other methods during the training process. By avoiding more complex solutions, we are able to keep the number of trainable parameters smaller, thereby keeping our training and sampling procedures as computationally efficient as possible. As a result, our generated images exhibit no discernible color shifts even when we generate samples with a simple Euler-Maruyama sampler.  We plan on expanding on more of the details of this architectural choice in a separate study. 

\subsection{Downscaling Diffusion Bridges}
In order to carry out our diffusion-based downscaling method, we train a contextual diffusion model for our high resolution dataset. Since the forward noising process is independent of the score function, we do not need to train a model for the low resolution dataset (we would only need that if we wanted to generate low resolution samples as well). In other words, the noising direction acts like a pre-trained encoder does within an hierarchical autoencoder setup \citep{Luo2022}.

Details on the construction of the diffusion models, the network architecture, the loss function, the training procedures, and the sampling method are provided in Appendices \ref{sec:dm_background}, \ref{sec:contextual_diffusion}, \ref{sec:NetworkArch}, \ref{sec:mean_bypass}, \ref{sec:Training}, and \ref{sec:sampling}. We note here that our network architecture does not preserve the doubly-periodic nature of the flow fields. This is because this is a unique feature of this data set which will not be present in most applications, for example, in downscaling patches of a larger fluid simulation.

\section{Results}\label{sec:results}
 In this section, we assess the quality of the samples generated from our diffusion bridge according to several metrics. In terms of bias correction, we focus on biases in spatial mean values, intermediate scale biases, and biases in more extreme tail events (e.g., tails of distributions). We additionally quantify how well large scale information is retained, how well small scale information is added by our model, and explore how the model generalizes to an unseen contextual field.

\begin{figure*}[htbp]
\centering
\noindent\includegraphics[width=\textwidth]{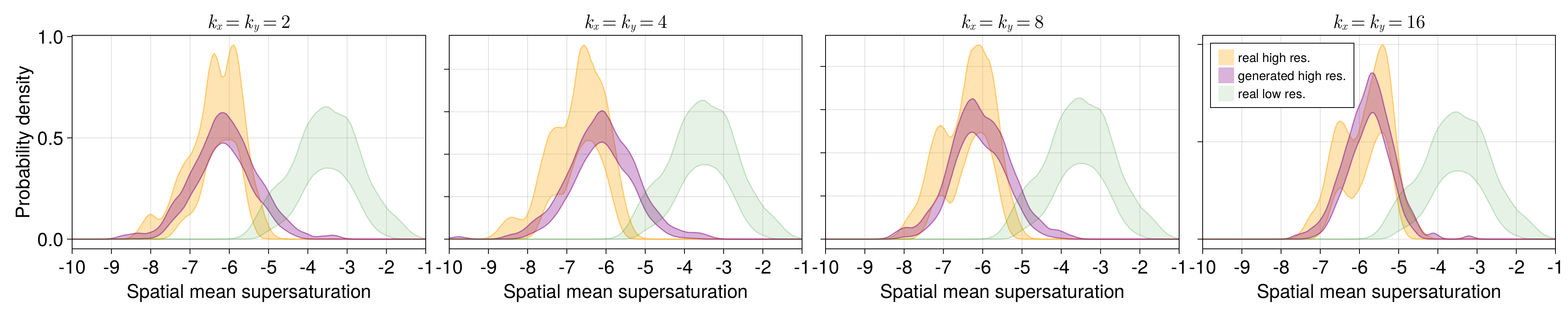}
\caption{Probability density function estimates for spatial means of the \textbf{supersaturation tracer} field. Columns correspond to different high resolution data subsets, where $k$ indicates the saturation specific humidity modulation wavenumber. Orange distributions show distributions of real high resolution samples, purple distributions show distributions of generated and downscaled high resolution samples, and green distributions show distributions of real low resolution samples. Shaded areas are computed from $10000$ bootstrap samples at the $99~\%$ confidence interval.}
\label{mean_ch1}
\end{figure*}
\begin{figure*}[htbp]
\centering
\noindent\includegraphics[width=\textwidth]{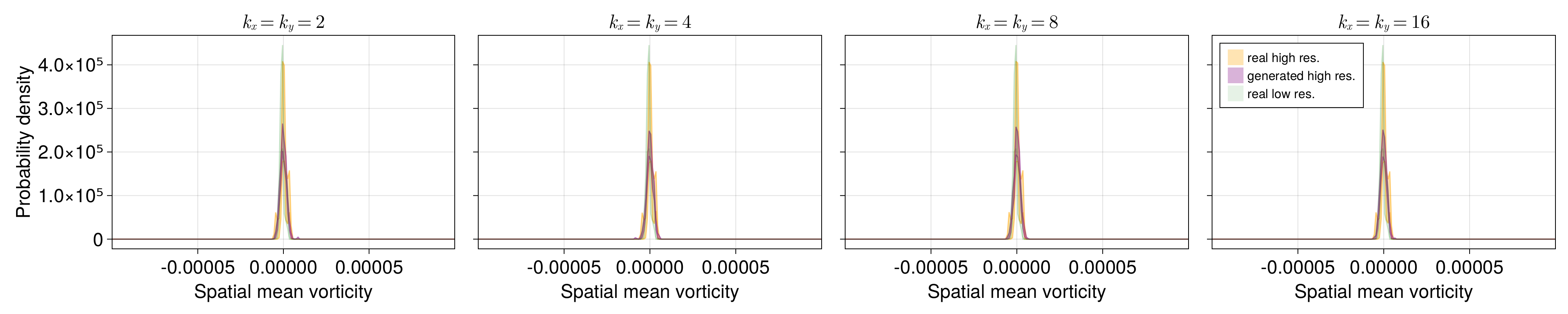}
\caption{Same as Figure~\ref{mean_ch1} but we are now plotting all quantities for \textbf{vorticity}. Spatial mean vorticity is not conserved in our simulation, but fluctuates around zero with small amplitude.}
\label{mean_ch2}
\end{figure*}

\subsection{Distributions of Supersaturation Tracer Vorticity Values}
Figures~\ref{pdfs_ch1}~\&~\ref{pdfs_ch2} show probability density functions for the supersaturation tracer and vorticity fields for four different context fields. The value of $k_x = k_y$ indicates the modulation wavenumber used in the context field, according to Equation~\eqref{eq:sat_moisture}. The green probability density functions show the distribution of values for the real low resolution data, while the orange probability density functions show the values for the real high resolution data. The purple probability density functions show the distribution of values from the downscaled (e.g., generated) high resolution samples using the context-dependent diffusion bridge approach. 

Figures~\ref{pdfs_ch1}\&~\ref{pdfs_ch2} show that the context-dependent diffusion bridge approach shifts the distribution of the field values computed from the low resolution dataset close to the distribution of values computed from the high resolution dataset. One can see in the both the mean and variance of the distribution is adjusted by downscaling so that the generated and high resolution distributions are much closer to each other than generated and low resolution distributions. Even though this is the case, the left tails remain consistently underestimated by an $\mathcal{O}(1)$ factor. We checked that random generated high resolution images ({\it not} downscaled low resolution images) demonstrated the same behavior (not shown), indicating that these errors originates in the model itself and not in the downscaling procedure.

\begin{figure*}[htbp]
\centering
\noindent\includegraphics[width=\textwidth]{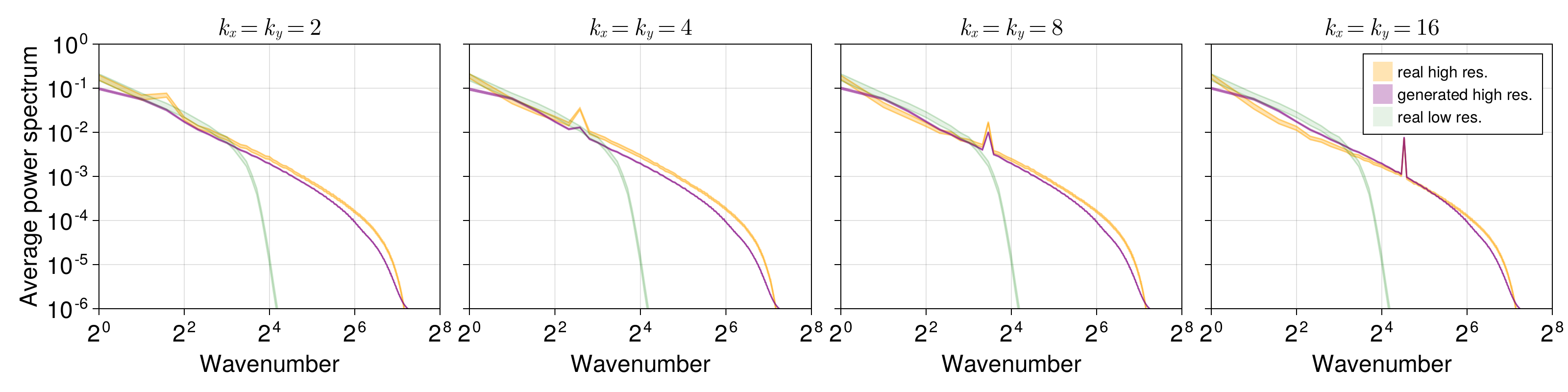}
\caption{Azimuthally averaged spectral density function estimates of \textbf{supersaturation tracer} field values. Columns correspond to different high resolution data subsets, where $k$ indicates the saturation specific humidity modulation wavenumber. Orange spectra show spectra of real high resolution samples, purple spectra show spectra of generated and downscaled high resolution samples, and green spectra show spectra of real low resolution samples. Shaded areas are computed from $10000$ bootstrap samples at the $99~\%$ confidence interval.}
\label{spectra_ch1}
\end{figure*}
\begin{figure*}[htbp]
\centering
\noindent\includegraphics[width=\textwidth]{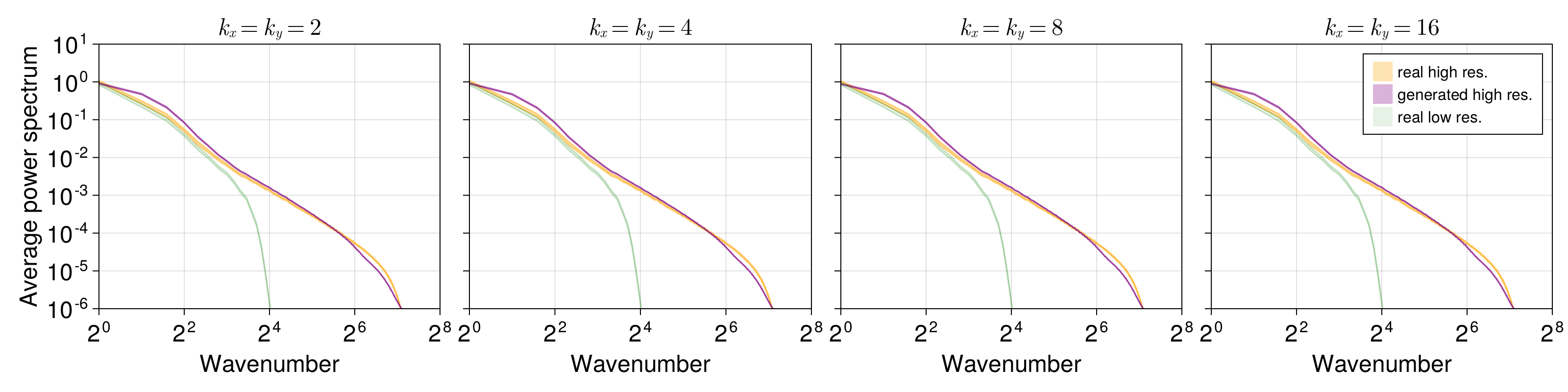}
\caption{Same as Figure~\ref{spectra_ch1} but we are now plotting all quantities for \textbf{vorticity} field values (e.g., the enstrophy spectrum).}
\label{spectra_ch2}
\end{figure*}

\subsection{Spatial Means of Supersaturation Tracer and Vorticity}
Figure~\ref{mean_ch1} shows the probability density function estimates for the spatial mean of the supersaturation tracer field. These demonstrate that the low resolution simulations differ from the high resolution simulations even in the spatial mean of the supersaturation tracer field. This indicates that the diffusion bridge-based downscaling approach does more than just adding in the small spatial scale features; it corrects biases in large-scale features as well. Although we find that the spatial mean biases get corrected, the variance of spatial mean values is larger in the generated data samples than it is for the real high resolution data. We hypothesize that a refinement of the mean-bypass layer could help alleviate this discrepancy.

Figure~\ref{mean_ch2} shows the probability density function estimates for spatial means of the vorticity field. By design of the two-dimensional fluid dynamics model, the spatial mean of the vorticity fluctuates around the zero and is nearly conserved. This is recovered in the generated data samples and is the result of our choice of including a mean-bypass layer in our neural network design, as described in Section \ref{sec:color_shift_main_text} and Appendix \ref{sec:mean_bypass}. Without the mean bypass layer in our modified U-Net architecture, we find that is is more difficult to obtain data samples with minimal spread in spatial mean vorticity.

Again, we checked that random generated high resolution images ({\it not} downscaled low resolution images) demonstrated the same behavior (not shown), indicating that these errors originates in the model itself and not in the downscaling procedure.

\subsection{Power Spectral Densities and the Role of Contextual Information}
Our fluid simulations depend on the contextual field via the supersaturation tracer equation, and the vorticity field is unaffected by this information. However, the role that the context plays in the high resolution supersaturation field is not clear from the distribution of pixel values and spatial means (Figures~\ref{pdfs_ch1} and \ref{mean_ch1}). Here we explore the role of contextual information via the power spectral density of the flow. This metric also allows us to understand how well the downscaling method fills in information on small scales and corrects intermediate biases.

Figures~\ref{spectra_ch1}~\&~\ref{spectra_ch2} show how the context-dependent diffusion bridge downscaling algorithm performs in spectral space. Figure~\ref{spectra_ch1} shows the mean azimuthally averaged power spectral density for the supersaturation tracer field and Figure~\ref{spectra_ch2} shows the mean azimuthally averaged power spectral density for the vorticity field. The green spectra show the distribution of values for the real low resolution data, while the orange spectra show the values for the real high resolution data. The purple spectra show the distribution of values from the downscaled (i.e., generated) high resolution samples using the context-dependent diffusion bridge approach. One can see that the real low resolution spectra decay rapidly already at relatively low spatial wavenumbers. This is due to the increased damping of small scales in the fluid dynamical simulations. The context-dependent diffusion bridge approach not only ``fills in'' the missing part of the spectra when comparing low resolution and high resolution datasets, but also corrects the intermediate scale bias stemming from the contextual information, i.e. the modulation of the background saturation specific humidity field in the high resolution simulations.

For all wavenumbers, there is an overall lack of power at all scales of the generated images, compared with the real high-resolution images, though the overall shape is correct. This implies that the correct spatial patterns are being learned, but that the overall contrast of the generated images is slightly muted. This was observed during training; we speculate that a more refined neural network architecture for the score combined with more data would alleviate this disagreement. Again, this same behavior was seen in random generated images and is therefore due to the model itself, and not to the downscaling procedure.

More importantly, the generated images at wavenumbers $k_{x,y}=4$ and $k_{x,y}=2$ are lacking power at the wavenumber that the context imposes on the flow. This was also observed in Figure~\ref{downscale_context}, and discussed in Section~\ref{sec:context_main_text}. It is because our algorithm requires choosing a value of $t^\star$ which balances preserving the large scale features of the flow with adding in the intermediate and small scale features. For the larger wavenumber contexts, this balance does not exist. That is, if the high resolution and low resolution fluid flows differ on essentially all scales, our method cannot work. Whether or not this type of translation would still be ``downscaling" is unclear. We verified that this lack of power at the contextual wavenumbers in the generated images was not due to the model itself; the purely generative high resolution model (using $t^\star = 1$) does correctly add in the contextual features in all cases; such a value of $t^\star$ would completely lose the low resolution features we are trying to preserve in downscaling. 

\begin{figure}[htbp]
\centering
\includegraphics[width=0.5\textwidth]{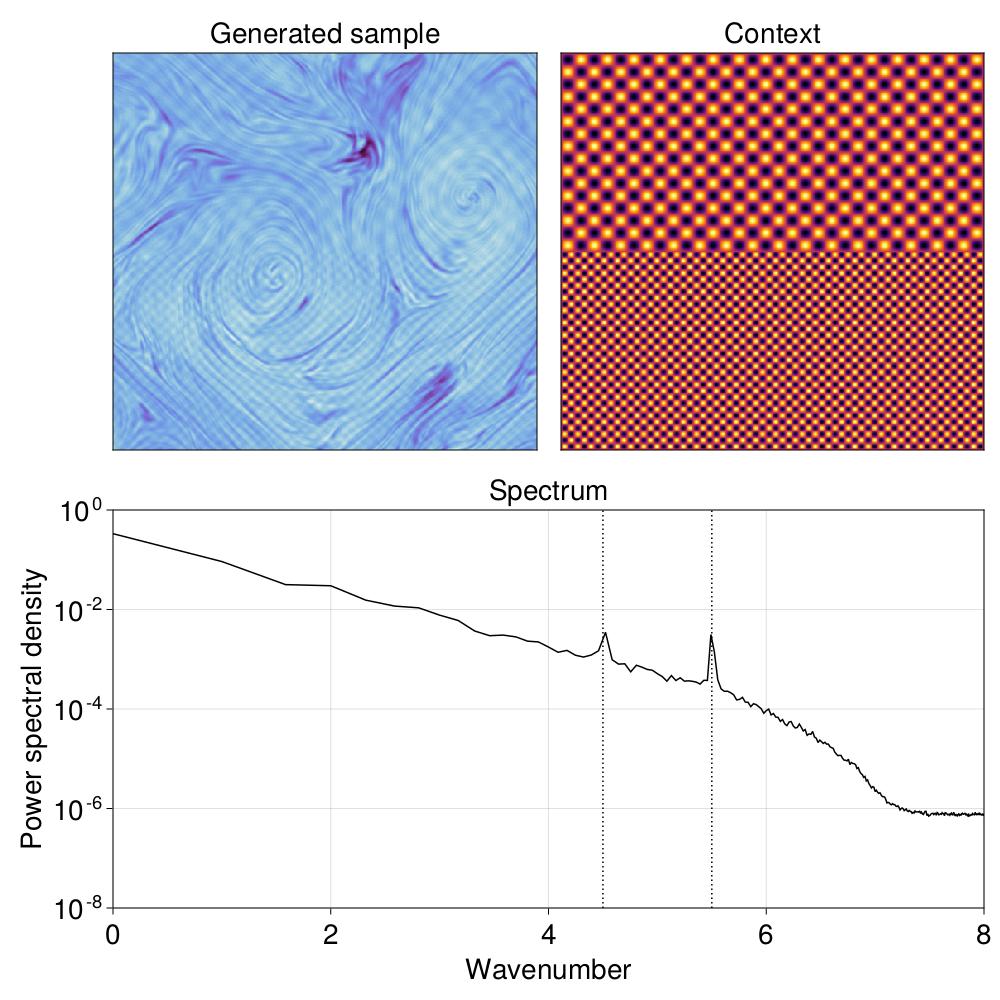}
\caption{Demonstration of the diffusion bridge algorithm using a contextual field that was not seen during training. A downscaled \textbf{supersaturation tracer} field is shown on the left, the context used during sampling is presented in the right panel, and the power spectral density of the downscaled image is shown on the bottom. The dashed lines indicate the wavenumber of the two contextual spatial frequencies.}
\label{superpose}
\end{figure}

These figures also demonstrate the role that context plays in the downscaling procedure, as it clearly affects the power spectrum of the resulting images. We took an initial step of testing our contextual diffusion model on an unseen context. To assess this, we created a new context using two wavenumbers, one of which was not used during training, and carried out the downscaling algorithm. The generated image, context field, and power spectral density of the supersaturation tracer are shown in Figure \ref{superpose}. The model correctly imposes the modulation, though it is hard to quantify its performance more quantitatively given that we do not have real fluid simulations with this context.

\subsection{Condensation Rate Distributions}
\begin{figure*}[htbp]
\centering
\noindent\includegraphics[width=\textwidth]{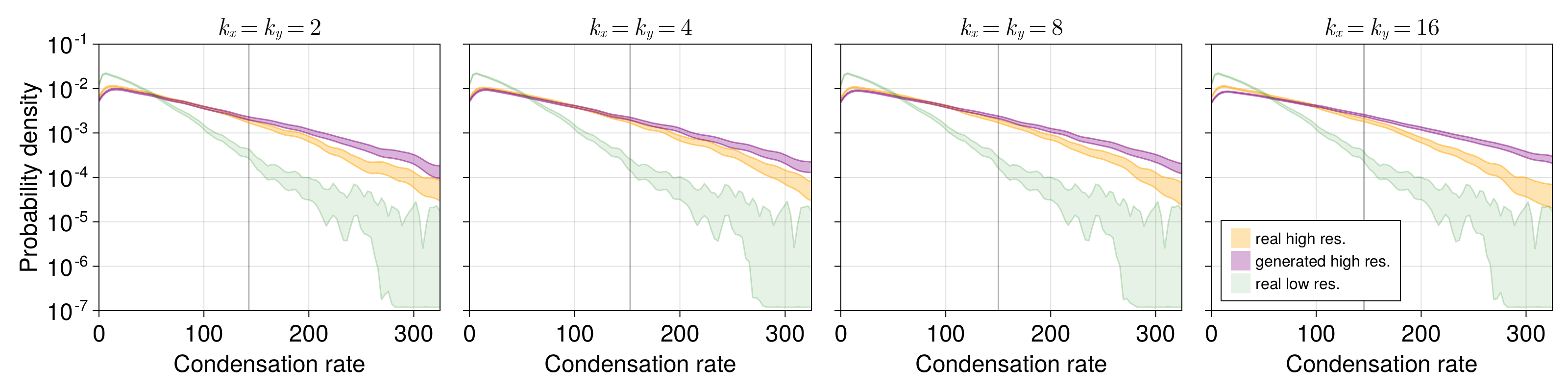}
\caption{Probability density function estimates for instantaneous \textbf{condensation rate}. Columns correspond to different high resolution data subsets, where $k$ indicates the saturation specific humidity modulation wavenumber. Orange distributions show distributions of real high resolution samples, purple distributions show distributions of generated and downscaled high resolution samples, and green distributions show distributions of real low resolution samples. Shaded areas are computed from $10000$ bootstrap samples at the $99~\%$ confidence interval. The vertical line approximately indicates the $90$-th percentile of true high resolution condensation rates.}
\label{cond_rate}
\end{figure*}

In order to further assess the performance of our downscaling method, we compute the distribution of the condensation rate for low and high resolution datasets. In order to do this, we calculate a kernel density estimate of positive condensation rates over the data. The calculation of the condensation rate is given in Appendix~\ref{sec:FluidModel} and can be thought of as a rain formation rate in the idealized model. Figure~\ref{cond_rate} shows how the downscaling algorithm performs when evaluating the distributions of condensation rate. The green distributions show the condensation rates for the low resolution data, while the orange distributions show the condensation rates for the real high resolution data. The purple distributions show the condensation rates from the downscaled (e.g., generated) high resolution samples using the context-dependent diffusion bridge approach. One can see that the tails of the high resolution data are underestimated by the low resolution distribution by one or two orders of magnitude, especially for very rare events. This is because sharp peaks are smoothed out in low resolution numerical simulations of fluid flows. The downscaling procedure ``lifts'' the tails up and alleviates the biases in condensation rate tail events. 

However, we find that the generated samples overestimate the occurrence of very rare events (e.g., $1/1000$ events). There are many reasons why this could be the case. In general, machine learning models may perform poorly in the tails of distributions due to the lack of training data from this part of the data domain. However, we speculate that apart from this general difficulty, diffusion models can also leave a small amount of residual noise in the generated samples that is imperceptible to the human eye, but that manifests itself in tail statistics. This is due to the specific choice of noising schedule of many diffusion models in which the final noise added during sample generation is not equal to zero. In order to make further improvements to this issue, it may be necessary to find an improved noising schedule. Due to numerical instabilities that can appear when the final noise amplitude approaches zero, we leave this technical challenge for future work. We also note that errors in the tail of a distribution can arise from errors in the means. Our generated images have larger variance in the means compared with the real images (Figure~\ref{mean_ch1}, which may also contribute to a shift in the tails. 

\subsection{Conditional Sampling Assessment}
Using the notation from Section \ref{sec:context_main_text}, we expect that a reasonable downscaling algorithm approximately generates samples from the conditional distribution $p(\vec{x}_\mathcal{T} |\vec{x}_\mathcal{S})$, where $\mathcal{T}$ is the high resolution data domain and $\mathcal{S}$ is the low resolution data domain. More concretely, we expect the large scale spatial features to be preserved between $\vec{x}_\mathcal{T}$ and $\vec{x}_\mathcal{S}$. To test how well our algorithm meets this requirement, we compute the distance between downscaled images and their low resolution source images using the pixel-wise $L^2$-metric. We additionally compute the same statistic for two randomly chosen low resolution images, and compare the distribution of these distances in each case to each other.

\begin{figure}[htbp]
\centering
\includegraphics[width=0.5\textwidth]{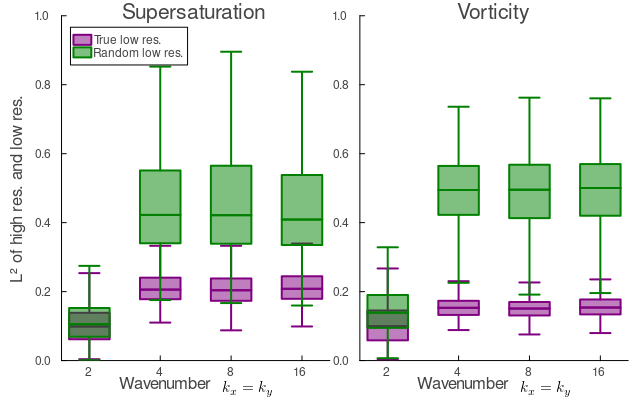}
\caption{Comparison of downscaled images, by channel and by wavenumber, with their low resolution source images. The statistics of the pixel-wise $L^2$-metric values between a filtered downscaled image and its filtered source (low resolution) image are shown in purple; the same statistics between two random low resolution images are shown in green.}
\label{l2_boxplot}
\end{figure}

The resulting distribution of $L^2$-metric values are shown via boxplot in Figure~\ref{l2_boxplot}. This demonstrates that the downscaled images are more similar to their low resolution source images than two randomly chosen low resolution images are to each other, indicating that broad spatial features are preserved by the diffusion bridge algorithm. Note that because biases exist between the high and low resolution data sets, as demonstrated in Figures~\ref{mean_ch1}, \ref{spectra_ch1}, and \ref{spectra_ch2}, we first carried out the following transformation before computing the distance metric. We low-pass filtered the images such that only spatial frequencies with $k< k^\star$ are present. We then normalized the images using the mean pixel value and standard deviation of the pixel values for the data domain in question. If we do not account for this, the $L^2$-metric value between the downscaled and real source images can be be very large, but mostly due to the biases of the low resolution data.

\section{Conclusions}
We have shown that a downscaling approach using context-dependent diffusion bridges can correct spatial mean biases and intermediate scale biases, as well as improve resolution in idealized low resolution fluid dynamics simulations. In addition, we showed that this approach can help to ``lift the tails'' of low resolution condensation rates, leading to at least an order of magnitude correction probabilites density values for the condensation rate tails. This suggests that diffusion-based generative models may be able to correct biases in extreme event rates even in more realistic settings and even without any explicit emphasis in the training loss function. We also demonstrated that the diffusion bridge method creates downscaled images which match the statistics of the azimuthally averaged power spectrum and distributions of supersaturation tracer and vorticity values of the original high-resolution data. By introducing a bypass connection in the neural network used to model the score in the reverse diffusion process within the diffusion bridge, the method alleviates the spatial mean bias (e.g., color shift) problem and preserves the value of the spatial mean vorticity. This implies that conservation laws based on global integrals may naturally be respected by diffusion models without further explicit emphasis in the loss function. While this may not be important for applications to realistic climate simulations, where smaller sections of the flow are downscaled one at a time, it may be useful in other contexts.

As pointed out in the introduction, diffusion models can have advantages over classical and other generative machine learning methods for downscaling. We find that their usefulness can be summarized as follows:
\begin{itemize}
\item Diffusion models are flexible and reusable. The downscaling approach developed and applied in this work did not require any special tuning for the datasets at hand and it did not require the low-resolution data during training at all. Domain translation tasks between data generated with other models or taken from observations only require training a diffusion model for each domain, thereby reducing the computational effort required during training. 
\item The loss functions used for training diffusion models in this work where generic and essentially unmodified and as such did not have any particular emphasis on extreme events. No quantile loss or spectral loss function was used in the training of our models. 
\item Diffusion bridges are able to approximately generate samples from high-resolution conditional distributions. This can be useful in applications scenarios where complex statistical quantities need to be computed or where it is not know what kind of statistical quantities need to be computed later on after training.
\end{itemize}
While some of these advantages may not apply in every modeling scenario, we find that overall the large flexibility of diffusion-based models makes them an appealing choice in generative modeling scenarios.

\subsection{Alternatives \& Future Directions}
The data used in this work is comprised of a two-dimensional forced turbulent fluid and a supersaturation tracer. It included several features which are similar to a more complex climate simulation (non-Gaussian statistics of the supersaturation tracer, the influence of site-specific orography-like features). However, an obvious next step is to test the approach presented in this work with a realistic climate data set and compare its performance more directly to other existing downscaling methods. In addition, we also identify some possible research directions and outstanding questions:
\begin{itemize}
    \item Downscaling with diffusion-based generative models could also be achieved without diffusion bridges. Instead one could use a projection step during sample generation similar to what was demonstrated in \citet{song2022solving}.
    \item Temporal coherence of samples may be achievable with diffusion models that are used in the context of video generation \citet{HoVideoDiffusion}. It would be interesting to test their performance on physical systems, but there may be drawbacks with respect to computational cost that need to be addressed.
    \item Guided sampling techniques for diffusion models, as introduced in \citet{ClassifierFreeGuidance}, may be useful in order to generate samples that have additional desirable characteristics, such as high values of certain climate indices, etc.
\end{itemize}
However, there is still room to extend the scope of the current work. We have already identified minor discrepancies in the downscaled images compared with the real high-resolution images, as discussed in Section \ref{sec:results}, and determined that these were largely due to the model itself (rather than the downscaling procedure). A study refining network architectures and the training procedure may improve the model and results. Additionally, using more varied contexts in training, and truly demonstrating generalization to out-of-sample contexts, is an important next step. 

Overall, it appears that diffusion-based models are promising candidates for future applications in the Earth sciences.

\clearpage
\acknowledgments
This research has been supported by Eric and Wendy Schmidt (by recommendations of the Schmidt Futures) and by the Cisco Foundation. The authors thank Simone Silvestri, Ricardo Baptista, Nikola Kovachki, Honkai Zheng, Andrew Stuart, and Tapio Schneider for stimulating discussions on this work. We thank Andre Souza, who assisted with an earlier version of the fluid dynamics model used in this study. Andre Souza and Tapio Schneider provided helpful comments on an early draft of this manuscript. We also acknowledge the authors of the fluid simulation code \hyperlink{https://github.com/FourierFlows/FourierFlows.jl}{FourierFlows.jl} which our simulations are based on. All numerical calculations and model training used for this manuscript where performed with the help of Caltech's Resnick High Performance Computing Center.

\datastatement
A Github repository with our training and analysis code for the diffusion model can be found at
\hyperlink{https://github.com/CliMA/diffusion-bridge-downscaling}{https://github.com/CliMA/diffusion-bridge-downscaling}. A Github repository for fluid simulations can be found at \hyperlink{https://github.com/CliMA/SimpleMoisture.jl}{https://github.com/CliMA/SimpleMoisture.jl}.

\appendix[A]
\appendixtitle{Data Generating Model}\label{sec:FluidModel}
The data generating model used in this work consists of a dynamical system that mimics the advection and condensation of moisture along isentropes Earth's extratropical atmosphere. It is idealized but contains enough complexity to test the performance of the machine learning algorithms outlined in this work. Specifically, it exhibits some desirable dynamical and statistical properties. For example, the supersaturation field is highly variable in space and the associated idealized instantaneous condensation rate follows a distribution with approximately exponential tails \citep{ogorman_2006}. These properties are useful when evaluating the skill of generative machine learning models in terms of spectral and statistical accuracy, especially with respect to extreme events.

The motivation for and behavior of this model are described extensively in \citet{ogorman_2006} and as a result, we only recapitulate the main ingredients of the model here. At its heart the model consists of the two-dimensional vorticity equation on a periodic domain forced randomly and damped via linear drag and hyperdiffusion (spectral filtering is an alternative). The governing equations of the vorticity field read
\begin{align}
\label{eq:vorticity}
    \partial_t \zeta + \partial_y \Psi \partial_x \zeta - \partial_x \Psi \partial_y \zeta &= f - a \zeta - \kappa \Delta^8 \zeta\\
    \Delta \Psi &= \zeta, 
\end{align}
where $\zeta$ denotes the vorticity, $\Psi$ is the stream function, $f$ is a stochastic forcing with an isotropic wavenumber spectrum and power contained in a narrow ring in wavenumber space centered on $k_f$ with bandwidth $\Delta k$, $a$ denotes a frictional time scale, and $\kappa$ acts as a hyperdiffusivity. The equation for the specific humidity field $q$ is given by
\begin{align}
\label{eq:moisture}
    \partial_t q + \partial_y \Psi \partial_x q - \partial_x \Psi \partial_y q &= e - c -\kappa \Delta^8 q\\
    c &= \frac{1}{\tau}(q-q_s) \Theta(q-q_s), 
\end{align}
where $\Psi$ is again the stream function, $e$ is an evaporation rate, taken as fixed in space and time, and $c$ denotes the instantaneous condensation rate.  Here, $\kappa$ is the same hyperdiffusivity as in Equation~\eqref{eq:vorticity}. The condensation rate $c$ is proportional to the difference between the specific humidity $q$ and the saturation specific humidity $q_s$, but condensation is only active when $q > q_s$, as it would be in Earth's atmosphere. In our simulations, we consider the case where the condensation time scale $\tau$ is small and finite. In other words, supersaturated ($q > q_s$) regions are relaxed back to the saturation specific humidity $q_s$ over a time scale $\tau$. The finiteness of $\tau$ mimics non-equilibrium thermodynamic processes but is not essential for the conclusions of this work. As described in the main text, we vary $q_s$ as a function of space to mimic both the decay of $q_s$ along isentropes in Earth's atmosphere and to impose spatial inhomogeneities at different lengths scales (Equation \eqref{eq:sat_moisture}). For large mean saturation deficits ($q < q_\mathrm{s}$), condensation events are rare, and the mean condensation rate tends to zero. For large evaporation rates, evaporation overpowers the ability of the turbulence to generate subsaturated fluid parcels through advection up the mean moisture gradient (cf. \citet{ogorman_2006}).
\begin{table*}[htbp]
\caption{Parameter values for data generating model. The complete dataset generated for this work consists of six subsets, a low-resolution dataset without any saturation specific humidity modulation, and five high-resolution subsets with varying modulation wavenumber. All simulations were run for $200,000$ time steps and the first $100,000$ time steps were discarded as spin-up for the purposes of the work presented here. The subset size reported includes the spin-up. All values are in non-dimensional form.}
\label{tab:parameters}
\begin{center}
\begin{tabular}{llllll}
\topline
\textbf{Shared parameters} & & & & &\\
domain size ($L$) & time step ($\Delta t$) & time steps ($N_t$) & drag coefficient ($a$) & $q_s$-gradient ($\gamma$) & evaporation rate ($e$)\\
 $2\pi$ & 1e-3 & 100,000 & 1e-2 & 1.0 & 1.0 \\
relaxation time ($\tau$) & forcing wavenumber ($k_f$) & bandwidth ($\Delta k$) & energy input rate ($\epsilon$) & &\\
1e-2 & 3 & 2 & 0.1 & & \\
\midline\\
\textbf{Subset parameters} & & & & &\\
name & resolution (LxL) & amplitude ($A$) & wavenumber ($k_{x,y}$) & hyperdiffusivity ($\kappa$) & subset size ($N_d$)\\
low-res & 64x64 & 0 & no modulation & 1e-8 & 2000\\
high-res-1 & 512x512 & 1 & 1 & 1e-16 & 2000\\
high-res-2 & 512x512 & 1 & 2 & 1e-16 & 2000\\
high-res-4 & 512x512 & 1 & 4 & 1e-16 & 2000\\
high-res-8 & 512x512 & 1 & 8 & 1e-16 &2000\\
high-res-16 & 512x512 & 1 & 16 & 1e-16 & 2000\\
\botline
\end{tabular}
\end{center}
\end{table*}

The complete dataset consists of six subsets, one low-resolution subset, and five high-resolution subsets with a total of $12,000$ datapoints. A summary of the parameters used to generate the complete dataset is given in Table~\ref{tab:parameters}. 

\appendix[B]
\section{Diffusion Models}\label{sec:dm_background}
As described in the main text, our diffusion model's noising process adds Gaussian noise to the image at each timestep. We have adopted the so-called ``variance-exploding" schedule where
\begin{subequations}
\label{eq:noise_schedule}
\begin{align}
g(t) &= \sigma_\mathrm{min}\bigg(\frac{\sigma_\mathrm{max}}{\sigma_\mathrm{min}} \bigg)^t \sqrt{2 \log{\bigg(\frac{\sigma_\mathrm{max}}{\sigma_\mathrm{min}} \bigg)}}\\
\sigma^2(t) & = \sigma_\mathrm{min}^2\bigg[\bigg(\frac{\sigma_\mathrm{max}}{\sigma_\mathrm{min}} \bigg)^{2t} - 1 \bigg] \approx \sigma_\mathrm{min}^2\bigg(\frac{\sigma_\mathrm{max}}{\sigma_\mathrm{min}} \bigg)^{2t}, 
\end{align}
\end{subequations}
where $\sigma_\mathrm{min}$ and $\sigma_\mathrm{max}$ are scalar parameters determining the shape of the variance with time. Other noising processes, including in the coefficient space after projecting onto a set of basis functions (\citet{Phillips2022}) and via a blurring process (\citet{Rissanen2022,Hoogeboom2022}), have also been used, but do not change the core idea of the diffusion model. 

Diffusion modeling parameterizes the score function $\vec{s}_\theta(\vec{x}, t) \approx s(\vec{x}, t)$, and optimizes the parameters through gradient descent on an appropriately chosen loss function. In practice, one usually represents the score function $\vec{s}_\vec{\theta}$ with a neural network $f_\theta$ defined by
\begin{equation}
\vec{s}_\theta(\vec{x},t) = \frac{f_\theta(\vec{x},t)}{\sigma(t)} \approx \vec{s}(\vec{x}, t).
\end{equation}
The benefit of this is that the neural network output will always be of $O(1)$, which can lead to an easier training task for the neural network (as opposed to forcing it to learn the prescribed $\sigma(t)$ dependence as well). The downside is that $\sigma(t=0)$ should ideally be zero, and as a result this introduces a singularity at $\sigma(t=0)$. In order to avoid this, it is standard practice \citep{Song2020_ImprovedTechniques} to instead set $\sigma(t=0) = \sigma_\mathrm{min}$, as given in the approximation of the expression for $\sigma^2(t)$ given by Equation \eqref{eq:noise_schedule}.

The denoising score-matching loss function \citep{Ho2020, ScoreBasedSDE_Song} is given by

\begin{subequations}
\label{eq:loss}
\begin{align}
\mathcal{L}(\vec{\theta}) &=  \mathbb{E}_{t, \vec{x}(0), \vec{x}(t)}\bigg[ \lambda(t)^2\bigg( \frac{\vec{f}_\vec{\theta}(\vec{x},t)}{\sigma(t)} - \nabla_\vec{x} \log{p(\vec{x}(t) | \vec{x}(0))}\bigg)^2\bigg] \\
& = \mathbb{E}_{t, \vec{x}(0), \vec{x}(t)} \bigg[\lambda(t)^2\bigg( \frac{\vec{f}_\vec{\theta}(\vec{x},t)}{\sigma(t)} - \frac{(\vec{x}(t) - \vec{x}(0))}{\sigma^2(t)}\bigg)^2  \bigg]\\
& = \mathbb{E}_{t, \vec{x}(0), \vec{x}(t)} \bigg[\frac{\lambda(t)^2}{\sigma(t)^2}\bigg( \vec{f}_\vec{\theta}(\vec{x},t) - \vec{\epsilon} \bigg)^2  \bigg],
\end{align}
\end{subequations}
where 
\begin{equation}
\mathbb{E}_{t, \vec{x}(0), \vec{x}(t)} =  \mathbb{E}_{t\sim U(0,1], \vec{x}(0) \sim p(\vec{x}(0)), \vec{x}(t) \sim p(\vec{x}(t) | \vec{x}(0)) },
\end{equation}
and $\vec{x}(t) = \vec{x}(0) + \sigma(t) \vec{\epsilon}$ is a noised image at time $t$, $\vec{\epsilon} \sim \mathcal{N}(\vec{0},\vec{1})$ is a Gaussian random vector, and $\lambda(t)$ is a weighting factor taken to be equal to $\sigma(t)$ (see \citet{song2021maximum}). From the last step in Equation~\eqref{eq:loss} one can see that the score matching loss is equivalent to making the neural net learn the added noise at time $t$. Note that although this involves an $L_2$ loss is between the score function and the gradient of the logarithm of the conditional distribution, optimizing $\mathcal{L}(\theta)$ results in an approximation to the true score of the unconditional distribution (\citet{Vincent2011}).

We slightly modified the above loss function to monitor the specific loss values with respect to spatial means and variations about the mean. As $\vec{\epsilon}$ in Equations \eqref{eq:loss} is random Gaussian noise, the mean $\bar{\vec{\epsilon}}$ is independent of the variations about the mean, $\vec{\epsilon}' = \vec{\epsilon} - \bar{\vec{\epsilon}}$. Since $\vec{f}_\theta$ seeks to match $\vec{\epsilon}$, we anticipate that the same will be true for it once the network is well-trained. In that case, we can rewrite the loss as

\begin{align}
\label{eq:loss_2term}
\mathcal{L}(\vec{\theta}) &= \mathbb{E}_{t, \vec{x}(0), \vec{x}(t)} \bigg[\frac{\lambda(t)^2}{\sigma(t)^2}\bigg( \vec{f}_\vec{\theta}(\vec{x},t) - \vec{\epsilon} \bigg)^2  \bigg] \\
&\approx \mathbb{E}_{t, \vec{x}(0), \vec{x}(t)} \bigg[\frac{\lambda(t)^2}{\sigma(t)^2}\bigg( (\vec{f}'_\vec{\theta}(\vec{x},t) - \vec{\epsilon}')^2 + (\bar{\vec{f}}_\vec{\theta}(\vec{x},t) - \bar{\vec{\epsilon}})^2 \bigg)  \bigg].
\end{align}

In practice this should not affect the training procedure; we found it useful mainly to track errors in the means which result in color shifts.

\section{Contextual Diffusion Models}\label{sec:contextual_diffusion}
When conditioning sampling on contextual information in climate modeling scenarios, such as topography, bathymetry, or land surface properties, we do have paired data points for high and low resolution images. Denoting these contextual images as $\vec{x}_\mathcal{C}$, we have access to samples from the joint distributions
\begin{subequations}
\begin{align}
\vec{z}_\mathcal{S} &= (\vec{x}_\mathcal{S}, \vec{x}_\mathcal{C}) \sim p(\vec{z}_\mathcal{S})\\
\vec{z}_\mathcal{T} &= (\vec{x}_\mathcal{T}, \vec{x}_\mathcal{C}) \sim p(\vec{z}_\mathcal{T}),
\end{align}
\end{subequations}
where, as in the main text, $\mathcal{T}$ and $\mathcal{S}$ denote the target and source domains. Then, we can follow \citet{ScoreBasedSDE_Song} to allow for conditional sampling. By optimizing the loss function

\begin{equation}\label{eq:conditional_loss}
\mathcal{L}(\vec{\theta}) = \mathbb{E}_{t, \vec{z}(0), \vec{z}(t) } \bigg[ \lambda(t)^2\bigg( \frac{\vec{f}_\vec{\theta}(\vec{z},t)}{\sigma(t)} - \nabla_\vec{x} \log{p(\vec{z}(t) | \vec{z}(0))}\bigg)^2\bigg],
\end{equation}
where
\begin{equation}
    \mathbb{E}_{t, \vec{z}(0), \vec{z}(t) } = \mathbb{E}_{t\sim U(0,1], \vec{z}(0) \sim p(\vec{z}(0)), \vec{z}(t) \sim p(\vec{z}(t) | \vec{z}(0)) },
\end{equation}
and $\vec{z}(t) = (\vec{x}(t), \vec{x}_\mathcal{C}(t))$ is the tuple containing the state of the fluid flow or climate model $\vec{x}(t)$ and the corresponding contextual information $\vec{x}_\mathcal{C}(t)$. We choose not to noise the context variables so that 
\begin{align}
p(\vec{z}(t) | \vec{z}(0)) = p(\vec{x}(t) | \vec{x}(0)) \delta(\vec{x}_\mathcal{C}(t) - \vec{x}_\mathcal{C}(0)),
\end{align}
and hence the score functions can be related as
\begin{equation}
\nabla_\vec{x} \log{p(\vec{z}(t) | \vec{z}(0))} = \nabla_\vec{x} \log{p(\vec{x}(t) | \vec{x}(0))}.
\end{equation}
The resulting score function in this contextual setup is then a known function, just like in the case of unconditional diffusion models. As shown in \citet{Batzolis2021}, optimizing this loss function is equivalent to learning a function $\vec{f}_\vec{\theta}(\vec{z},t) = \nabla_\vec{x} \log{p(\vec{x}(t) | \vec{x}_\mathcal{C})}$, i.e., one that represents the conditional score. In implementation, we realize this by inputting $\vec{x}_\mathcal{C}$ as an additional channel of the diffusion model input. More discussion of the architecture is given in Section \ref{sec:NetworkArch}.

\appendix[C]
\section{Network Architecture}\label{sec:NetworkArch}
The foundation of our score network is a U-net \citep{unet}, which maps two inputs ($\vec{X}$, a tensor of size $(N, N, C_{in}, B)$, and $t$, a tensor of size $(B,)$), to a single output $\vec{Y}$, a tensor of size $(N, N, C_{out}, B)$. That is, U-net returns
\begin{equation}
\vec{Y} = \mathcal{U}(\vec{X}, t;\vec{\theta}),
\end{equation}
where $\mathcal{U}$ denotes the U-net with parameters $\vec{\theta}$ described in more detail below.

The first input $\vec{X}$ holds a batch of images, the second $t$ is a batch of times; $B$ is the size of the batch. Any individual channel of the input or output is an image of size $(N,N)$; there are $C_{in}$ input channels and $C_{out}$ output channels. For our data set, our input images have two noised channels: the fluid vorticity and supersaturation tracer concentration. Including the contextual information, we have $C_{in} = 3, C_{out} = 2$. 

In our default configuration, the U-net has five distinct parts. The first is an initial lifting layer, which is a convolution that preserves the spatial dimensionality of $\vec{X}$, but increases the number of channels from $C_{in}$ to 32. Three downsampling (convolutional) layers follow, which reduce the spatial dimensionality by a factor of 2, and which increase the number of channels by a factor of 2. These transformed data are passed through eight residual blocks which preserve the dimensionality of the transformed data \citep{resnet}. Then, three upsampling layers, comprised of nearest neighbor upsampling, followed by convolutions, increase the spatial dimensionality while decreasing the number of channels, mirror the downsampling layers. Finally, a projection layer decreases the number of channels to $C_{out}$. We use $3$x$3$ convolutional kernels, group normalization \citep{wu2018group}, and the swish function as a nonlinearity \citep{swish}. 

The time variable is first embedded using a random Fourier projection \citep{tancik2020fourier}. This embedded time is then transformed by a dense network at each up- and down-sampling layer, after which it is the added to the up- or down-sampled image. The sum is then group normalized, and operated on by the swish function, following \citet{ScoreBasedSDE_Song, Ho2020}, before being passed to the next layer.

\section{Modifications to U-net: Mean Bypass Network}\label{sec:mean_bypass}
We modified the neural network architecture introduced in Sec.~\ref{sec:NetworkArch} to include a bypass connection. The incoming batch is split into a component that has spatial variations (in fact, the original image after subtracting the channel-and-batch-wise spatial means) and the spatial average of each channel and batch member. The spatial average is then fed through the bypass network, while the spatially varying component is fed through the U-net. At the final layer the output from the U-net and the bypass are added together, after removing the spatial average of the output of the U-net. In this way, we have a completely separate network handling the spatial means and the spatial variations about the mean.  Along with our choice of loss function, this has the added advantage of making the mean prediction and spatial variation prediction entirely independent tasks.

Concretely, we  compute the spatial mean, by channel and batch member, of the input $\vec{X}$, denoted $\bar{\vec{X}}$. The spatial variation about the mean is denoted as $\vec{X}' = \vec{X} - \bar{\vec{X}}$, and is processed by the U-net as discussed in Sec.~\ref{sec:NetworkArch}, to produce an output $\mathcal{U}(\vec{X}', t;\vec{\theta})$. We then subtract the spatial mean (by channel and batch member), from this output, i.e. we produce a tensor $\vec{Y}' = \mathcal{U}(\vec{X}', t;\vec{\theta}) - \bar{\mathcal{U}}(\vec{X}', t\theta)$ that has zero spatial mean.  A separate function $\mathcal{M}(\bar{\vec{X}}, t;\vec{\phi})$, with trainable parameters $\vec{\phi}$ operates on $\bar{\vec{X}}$ and $t$, and returns a tensor of the same size as $\bar{\vec{X}}$, denoted $\bar{\vec{Y}}$. In the last layer, we combine the outputs of these individual components to produce the final output $\vec{Y}$ as
\begin{equation}
\vec{Y} = \mathcal{U}(\vec{X}', t;\vec{\theta}) - \bar{\mathcal{U}}(\vec{X}', t;\theta) + \mathcal{M}(\bar{\vec{X}}, t;\vec{\phi}) = \vec{Y}' + \bar{\vec{Y}}. 
\end{equation}

The network $\mathcal{M}$ consists of a three-layer dense feed-forward network, consisting of two linear transformations followed by a normalization and nonlinear activation function, and a single final linear transformation, without an activation or normalization. The embedded time is handled in the exact same way as for $\mathcal{U}$; it is passed through a linear transformation before being added to the transformed input, prior to normalization and activation.

Note that because of this, our implemented solution does not take advantage of correlations between the spatial variations about the mean and the mean.  If spatial variations of the input are useful for predicting the spatial mean of the score, or vice versa, our prediction will not make use of that information.  Through limited testing, we found that letting $\mathcal{U}$ have access to the entire input $\vec{X}$ yielded slightly worse performance after training for the same number of epochs. More investigation is required in order to take into account these correlations.

\section{Model Training}\label{sec:Training}
We follow the recommendations of \citet{ScoreBasedSDE_Song} and \citet{Ho2020} in setting up the optimizer for score-matching denoising diffusion models. We use an Adam optimizer with a learning rate of $\lambda_0$ = 2e-4, $\epsilon=$1e-8, $\beta_1= 0.9$, $\beta_2$= 0.999. We employ gradient norm clipping to a value of 1.0. We additionally employ a linear warmup schedule in the learning rate, from 0 to $\lambda_0$, over 5000 gradient updates. A batchsize of 4 was used for all runs. We generally train for 125 epochs. In our tests, we found that a type of overfitting would occur if we ran for longer, and we used dropout in the residual layers, with a probability of 0.5, to help alleviate this.

With respect to preprocessing the raw images, we proceed as follows. We first split each data sample into a constant mean image and an image of deviations from the mean. Over these two components of the data, we carry out an independent minmax scaling, such that the minimum pixel value (over all of the preoprocessed data) is -1 and the maximum (over all of the preprocessed data) pixel value is 1. We then add the two back together. The resulting data set no longer has a minimum and maximum pixel value of exactly $\pm1$, but because the maximum and minimum values of the mean are not necessarily correlated with the images that have the maximum and minimum spatial deviations from the mean, the distribution of pixels is still mostly contained within the $[-1,1]$ range (and at worst, in the $[-2,2]$ range). 

Our main motivation for this preprocessing step is because the total vorticity is conserved, and so the distribution of the total vorticity is a delta function. Floating point error turns this into a Gaussian with a very small variance. Our preprocessing step then turns this into a much wider distribution, which will be easier to learn. However we expect that this is beneficial in general given that means are handled by an independent neural network; this is akin to preprocessing the input of that network as is standard practice.

\section{Sample Generation}\label{sec:sampling}
To generate all of the results shown here, we use the Euler-Maruyama (EM) method with a fixed timestep for solving the stochastic differential equations.  For the SDE
\begin{equation}
\mathrm{d}x = f(x,t) \mathrm{d}t + g(x,t) \mathrm{d}W,
\end{equation}
the update rule is as follows:
\begin{equation}
    x(t+\Delta t) = x(t) + f(x,t) \Delta t + g(x,t) \eta \sqrt{\Delta t},
\end{equation}
where $\eta \sim \mathcal{N}(0,1)$. For all simulations, we use a fixed timestep of 0.002, which corresponds to 500 steps from $t=\epsilon=1$e-5 to $t=1$. Image generation was not the dominant computational cost for this project, so we did not explore varying the time-stepping algorithm or timestep. Testing alternate time-stepping schemes is an activate area of research in the field.

\appendix[F] 
\appendixtitle{Azimuthally-averaged Power Spectral Density}\label{sec:PSD}
For each channel in the input data, we have a two dimensional image of dimensions $N\times N$. We compute the discrete Fourier transform of the image,
\begin{equation}
\tilde{I}(k_x, k_y) = \sum_{x=-N/2}^{N/2-1} \sum_{y=-N/2}^{N/2-1} I(x,y) \exp{[-i 2\pi/N(k_x x+k_y y)]}.
\end{equation} 
The power spectrum for wavenumbers $(k_x, k_y)$ is given by
\begin{equation}
    \mathrm{PS}(k_x, k_y) = \frac{1}{N^4} \tilde{I}(k_x, k_y)\tilde{I}^*(k_x, k_y).
\end{equation}
This can be converted into a power spectral density $\mathrm{PSD}(k_x, k_y)$ by dividing by an area in wavenumber space \citep{youngworth2005overview}. We may convert to polar coordinates ($k, \phi$), where $k = \sqrt{k_x^2+k_y^2}$. For isotropic flows,  the expectation of $|\tilde{I}(k, \phi)|$ over different regions of the flow is independent of $\phi$. This means carrying out an integral of the azimuthal angle $\phi$ leads to no loss of information (in expectation). We can write the azimuthally averaged power spectral density as
\begin{align}\label{eq:PSD}
\mathrm{PSD}(k) &= \frac{1}{N^4}\frac{\int_0^{2\pi}\int_k^{k+1} \tilde{I}^*\tilde{I} k' dk' d\phi}{\int_0^{2\pi} \int_k^{k+1}k' dk' d\phi}  \\
&\approx \frac{1}{N^4}\frac{\sum_{k_x}\sum_{k_y} \tilde{I}^*\tilde{I} \Theta(k^2 \leq k_x^2+k_y^2 < (k+1)^2)}{\sum_{k_x}\sum_{k_y}  \Theta(k^2 \leq k_x^2+k_y^2 < (k+1)^2)},
\end{align}
where $\Theta(\mathrm{condition})$ is a function which returns 1 when the condition is true and 0 otherwise. This is metric becomes less informative for images of flows with preferred directions or inhomogeneities, in which case the 2D Fourier transformed image itself may be more useful. 

Our Algorithm \ref{alg:algorithm1} requires knowing the power spectrum for white noise. One can show that if $I(x,y) \sim \mathcal{N}(0,\sigma^2)$, that $|\tilde{I}(k_x,k_y)^2| \sim \mathrm{Exp}[1/(\sigma^2 N^2)]$ when $k_x$ or $k_y$ are greater than zero. This has an expected value of $\sigma^2 N^2$. Plugging this into Equation \eqref{eq:PSD}, we see that the $\mathrm{PSD}(k)$ of Gaussian white noise is independent of wavenumber $k$ and has an expected value of $\sigma^2/N^2$ for $k>0$\footnote{When $k_x=k_y=0, |\tilde{I}(0,0)^2|$ is not drawn from an exponential distribution. Instead, $|\tilde{I}(0,0)^2|/(\sigma^2N^2)\sim \chi_1^2$, the chi-squared distribution. We subtract the means prior to computing the PSD, so $|\tilde{I}(0,0)^2|\approx 0$.}. 

\newpage
\bibliographystyle{ametsocV6}
\bibliography{references}

\end{document}